\documentclass[letterpaper, 10 pt, conference]{ieeeconf}  
\usepackage{float}
\usepackage{hyperref}
\IEEEoverridecommandlockouts                              
\usepackage{graphics} 
\usepackage{epsfig} 
\usepackage{mathptmx} 
\usepackage{times,color} 
\usepackage{amsmath} 
\usepackage{amssymb}  
\usepackage[sort,compress]{cite}  

\usepackage{todonotes}

\usepackage{caption,balance}
\usepackage[font=small]{caption}
\usepackage{array}
\usepackage{subcaption}
\newcolumntype{P}[1]{>{\centering\arraybackslash}p{#1}}
\newcolumntype{M}[1]{>{\centering\arraybackslash}m{#1}}

\newcommand{\argmax}{ \operatorname*{arg\,max}}

\newcommand{\squishlist}
{ \begin{list}{$\bullet$}
{ \setlength{\itemsep}{5pt}
  \setlength{\parsep}{0pt}
  \setlength{\topsep}{0pt}
  \setlength{\partopsep}{0pt}
  \setlength{\leftmargin}{0.7em}
  \setlength{\labelwidth}{0.5em}
  \setlength{\labelsep}{0.2em} 
} 
}

\newcommand{\squishlisttwo}
{ \begin{list}{$\bullet$}
{ \setlength{\itemsep}{1pt}
  \setlength{\parsep}{0pt}
  \setlength{\topsep}{1pt}
  \setlength{\partopsep}{0pt}
  \setlength{\leftmargin}{1em}
  \setlength{\labelwidth}{0.5em}
  \setlength{\labelsep}{0.5em}
} 
}

\newcommand{\squishend}{\end{list}}
  
\title{\LARGE \bf
Near-Optimal Adversarial Policy Switching for \\Decentralized Asynchronous Multi-Agent Systems
}

\def\myproof{1} 

\author{Trong Nghia Hoang$^{1}$, Yuchen Xiao$^{2}$, Kavinayan Sivakumar$^{3}$, Christopher Amato$^{2}$, and Jonathan P.\ How$^{1}$
\thanks{$^{1}$Trong Nghia Hoang and Jonathan How are with the Laboratory for Information and Decision Systems (LIDS), MIT, Cambridge, MA 02139, USA
        {\tt\small \{nghiaht,jhow\}@mit.edu}}%
\thanks{$^{2}$Yuchen Xiao and Christopher Amato are with the College of Computer and Information Science (CCIS), Northeastern University,
        Boston, MA 02115, USA
        {\tt\small \{xiao.yuch,c.amato\}@northeastern.edu}}%
\thanks{$^{3}$Kavinayan Sivakumar is with the Department of Electrical Engineering, Princeton University, 
		Princeton, NJ 08540, USA
		{\tt\small ks16@princeton.edu}}%
}

\begin{document}

\maketitle
\thispagestyle{empty}
\pagestyle{empty}

\begin{abstract}
A key challenge in multi-robot and multi-agent systems is generating solutions that are robust to other self-interested or even adversarial parties who actively try to prevent the agents from achieving their goals. The practicality of existing works addressing this challenge is limited to only small-scale synchronous decision-making scenarios or a single agent planning its best response against a single adversary with fixed, procedurally characterized strategies. In contrast this paper considers a more realistic class of problems where a team of asynchronous agents with limited observation and communication capabilities need to compete against multiple strategic adversaries with changing strategies. This problem necessitates agents that can coordinate to detect changes in adversary strategies and plan the best response accordingly. Our approach first optimizes a set of \emph{stratagems} that represent these best responses. These optimized stratagems are then integrated into a unified policy that can detect and respond when the adversaries change their strategies. The near-optimality of the proposed framework is established theoretically as well as demonstrated empirically in simulation and hardware. 
\end{abstract}

\section{INTRODUCTION}

Multi-robot systems is a widely studied field, but the research is typically focused on a single team of cooperative, or self-interested, robots \cite{Rubenstein14,Werfel14,Shayegan17}. 
In contrast, many real-world domains consist of a team of robots that must complete tasks while competing against other adversarial robots. For instance, consider a team of UAVs tasked with surveying a scene or locating a secret base as well as an opposing team of UAVs tasked with preventing the secret base from being found. These adversarial scenarios require reasoning about not only completing the tasks designated to the team, but also considering what the adversarial robots may do to prevent their completion. 
In this paper, we study the general multi-robot decision-making problem with uncertainty in outcomes, sensors and communication, while incorporating multiple adversarial robots into this problem. Communication uncertainty and limitations further necessitates the design of decentralized agents that can coordinate with their teammates while anticipating changes in the adversary strategies using only their partial views of the world. This is for the first time all these forms of uncertainty as well as adversarial behavior have been considered in the same decision-theoretic planning framework\footnote{Previous works \cite{Murray03,Stone07Book} in the literature addressing this challenge have mostly focused on reactive frameworks or do not consider multiple adversarial robots into their frameworks.}. Furthermore, the size of the spaces for possible set of actions and coordination strategies for both the teammates and adversaries scale exponentially in the number of agents \cite{Doshi2005} and are typically much too large for an agent to reason about directly. Hence, a successful agent usually requires some form of high-level abstraction to reduce its effective planning space \cite{Shayegan15,Shayegan16,Amato14}. 


One approach is therefore to create a set of basic stratagems, which are best-responses to particular forms of adversarial behavior. The reasoning problem is then reduced to choosing among these basic stratagems in a given situation, thus significantly improving the scalability of planning. 
This approach can be achieved by anticipating in advance a small set of high-level tactics from which the adversaries can choose in any situations, that capture the diversity of their intentions. The task of the high-level planner then is to choose a response to the adversaries' current tactics and follow it until it is determined that they have changed their tactics and a new response is needed. This fits particularly well in asynchronous robotic planning scenarios: since each stratagem has different execution time, agent decision making is no longer synchronized as assumed in existing non-cooperative multi-agent frameworks \cite{Doshi2005,NghiaIJCAI13a,NghiaIJCAI13b}. 



The main contribution of this paper therefore focuses on the design of such a high-level planner, which can be decoupled into two separate tasks. The first task involves generating a set of basic \emph{stratagems} for a team of decentralized agents, each of which is optimized to work best against a particular tactic of the adversaries. This is formulated as a set of Macro-Action Decentralized Partially Observable Markov Decision Processes (MacDec-POMDPs) \cite{Amato14,IJRR17MacDec} that each characterize a cooperative scenario where a team of decentralized agents collaborate to maximize the team's expected performance while operating in a stationary environment simulated by a single tactic of the adversaries (Section~\ref{skill}). The stratagems can therefore be acquired by solving for a set of probabilistic policy controllers that maximize the expected total reward generated by the corresponding MacDec-POMDPs. Then, the second task is to integrate these specialized policy controllers into a unified policy controller that works best on average against the adversaries' switching tactics. This again can be achieved by optimizing the unified controller with respect to a series of MacDec-POMDPs (Section~\ref{fusion}) so that it can detect situation changes and switch opportunistically between these stratagems to respond effectively to the adversaries' new tactical choice. Interestingly, it can be shown that under a certain mild assumption, the result of this stratagem integration/fusion scheme appears to be near optimal with high probability as shown in Section~\ref{guarantee}. Finally, to empirically demonstrate the effectiveness of the proposed framework, experiments conducted for a robotic scenarios are presented in Section~\ref{experiment}, which show consistent results with our theoretical analysis. 
  
\section{BACKGROUND AND NOTATIONS}
\label{DecSPOMDP}
This section provides a short overview of MacDec-POMDPs \cite{Amato14,IJRR17MacDec} for decentralized multi-agent decision-making under uncertainty. Formally, a MacDec-POMDP is defined as a Dec-POMDP \cite{Bernstein2000,Book16} tuple $(\mathbb{I}, \mathbb{S}, \{\mathbb{A}_i\}_{i=1}^n, \{\mathbb{O}_i\}_{i=1}^n \mathbb{T}, \mathbb{Z}, \mathbb{R}, \gamma, b)$ augmented with a finite set of macro-actions, $\mathbb{M}_i$, for each agent, $i$, with $\mathbb{M} \triangleq \mathbb{M}_1 \cup \ldots \cup \mathbb{M}_n$ denote the set of joint macro-actions. Each macro-action is defined as a tuple $m_i = (\beta_{m_i}, I_{m_i}, \rho_{m_i})$ where $\beta_{m_i} : \mathbb{S} \rightarrow \{0, 1\}$ and $I_{m_i} \in \mathbb{S}_i$ are sets of rules that decide, respectively, the termination and eligibility to initiate of the corresponding macro-action $m_i$, while $\rho_{m_i}: \Theta_i \rightarrow \mathbb{A}_i$ denotes a low-level policy that maps agent $i$'s local histories $\theta_i \in \Theta_i$ to primitive actions $a_i \in \mathbb{A}_i$. Each agent will follow a chosen macro-action $m_i$ until its termination condition $\beta_{m_i}$ is met. Its stream of observations collected during the execution of $m_i$ is jointly defined as a macro-observation $\eta_i$. As such, each individual high-level policy $\pi_i: \zeta_i \rightarrow m_i$ of agent $i$ can then be characterized as a mapping from its history of macro-actions and -observations $\zeta_i \triangleq \{(m_{t-1}^i, \eta_t^i)\}_{t\geq1}$ to the next macro-action. Planning in Dec-POMDP therefore involves maximizing the following total expected reward with respect to the joint high-level policy $\pi = (\pi_1, \pi_2, \ldots, \pi_n)$:
\begin{eqnarray}
\pi^\ast &=& \argmax_{\pi} \mathbb{E}\left[\sum_{t=0}^{+\infty} \gamma^t\mathbb{R}(s_t, a_t) | b, \mathbb{M}, \pi\right] \ \label{eq:2}
\end{eqnarray}

Unlike Dec-POMDP's, the MacDec-POMDP formalism is naturally suitable for asynchronous multi-robot planning scenarios since it is not necessary for the macro-actions $m_i = (\beta_{m_i}, I_{m_i}, \rho_{m_i})$ to share the same execution time. In fact, from the perspective of an individual agent, the outcome of its selected macro-action (e.g., when it terminates) is non-deterministic as its termination rule may depend on the global state of the environment as well as the movements of the other parties, which are not observable to the agent. This makes optimizing $\pi$ via \eqref{eq:2} using traditional model-based dynamic programming techniques \cite{Szer2005,Seuken2007a,Seuken2007b,Boularias2008,Spaan2008,Spaan2011,Amato14} possible only if the probability distribution over the stochastic outcome of $\beta_{m_i}$, e.g., $p\left(\beta_{m_i}(s) = 0\ |\ s\right)$, is explicitly characterized. This is not trivial and does not scale well in complex decision problems with long planning horizon, vast state and action spaces. Alternatively, to sidestep this difficulty, it is also possible to parameterize and optimize $\pi$ directly via interaction with a black-box simulator\footnote{In many real-world scenarios, it is often easier to hand-code a simulator that captures the interaction rules between agents than learning probability models of their outcomes.} that implicitly encodes the probabilistic models of transition $\mathbb{T}$, observation $\mathbb{Z}$, reward $\mathbb{R}$ and termination rule $\beta_{m_i}$ \cite{Shayegan16}. This interestingly allows us to avoid modeling these probabilistic models directly and improve the scalability of solving MacDec-POMDPs. The specifics of this model-free approach are detailed in Section~\ref{skill} which serves as the building block of our adversarial multi-agent planning paradigm in Section~\ref{fusion}.

\section{GENERATING BASIC STRATAGEMS}
\label{skill}
This section assumes we have access to a set of black-box simulators preset by the domain expert to simulate accurately the adversaries' basic tactics, upon which more advanced strategies might be built. For example, in popular real-time strategy (RTS) games, a player can often anticipate in advance a small set of effective basic tactics from which the other competitors might choose in any situations. The decision making process of a player therefore comprises two parts. The first part focuses on formulating fundamental stratagems to counter the anticipated tactics of the adversaries and is addressed in the remaining of this section. The second part then is to integrate the resulting stratagems into a unified strategy that can detect changes in the adversaries' tactical choice and switch opportunistically between them in response to those changes (see Section~\ref{fusion}). 

\begin{figure}[t]
\includegraphics[width=8.6cm,height=4cm]{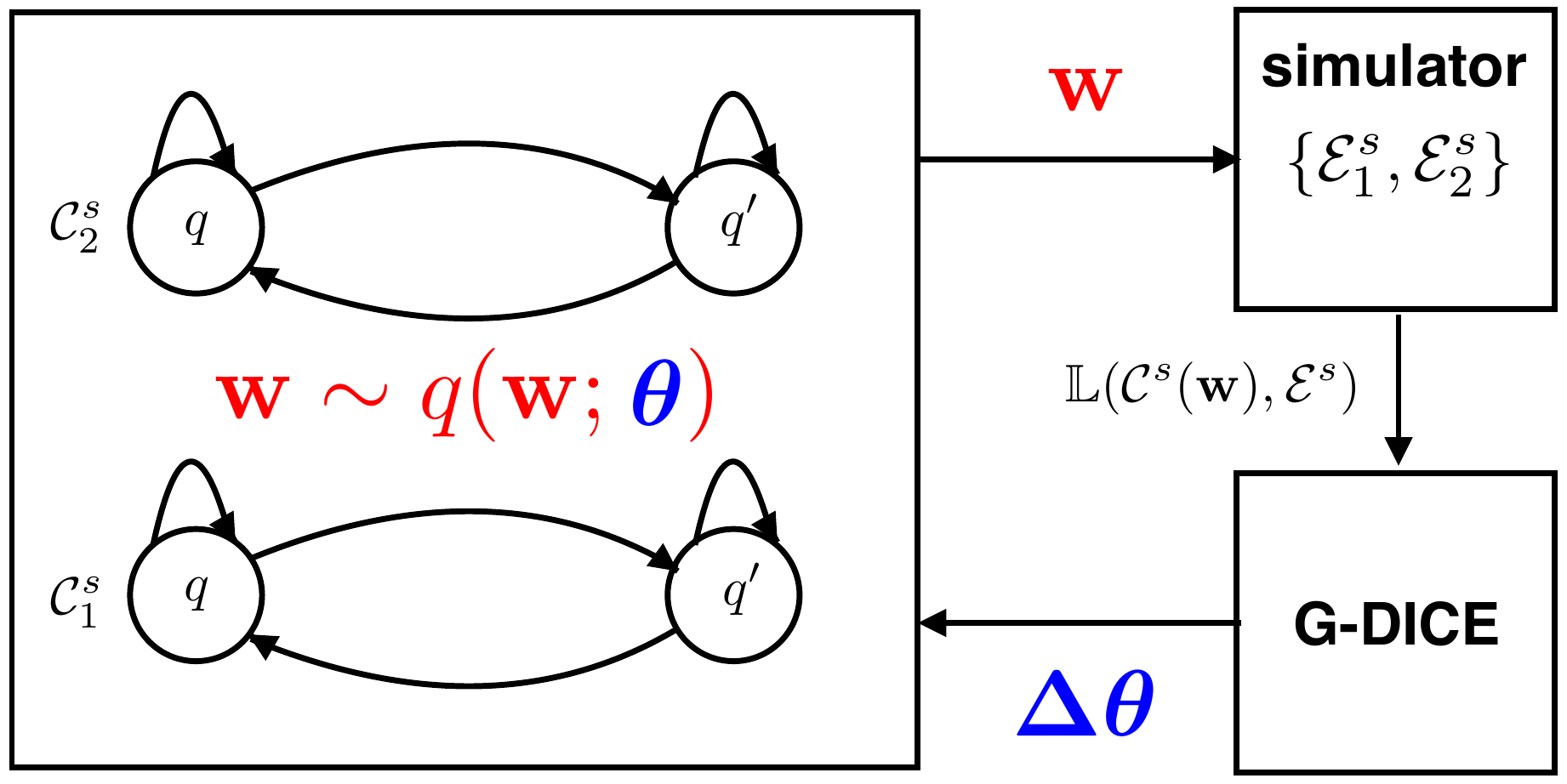} 
\caption{A team of two collaborative agents is represented by two decentralized controllers $\mathcal{C}^s \triangleq \{\mathcal{C}_1^s,\mathcal{C}^s_2\}$ characterizing their stratagem against a basic tactic $\mathcal{E}^s \triangleq \{\mathcal{E}^s_1, \mathcal{E}^s_2\}$ of the adversaries. The sampling distribution of $\mathcal{C}^s$'s parametric weights $\mathbf{w}$ is then optimized using performance feedback $\mathbb{L}(\mathcal{C}^s(\mathbf{w}),\mathcal{E}^s)$ from interacting with $\mathcal{E}^s$'s black-box simulator.}\vspace{-3mm}
\label{fig:1}
\end{figure}

In particular, formulating a stratagem to counter a specific tactic of the adversaries can be posed as solving a MacDec-POMDP which characterizes a cooperative scenario where a team of decentralized agents collaborate to maximize their total expected reward while operating in an artificial environment driven by the corresponding tactic simulator. The stratagem can then be optimized via simulation as detailed next. Formally, we represent a stratagem of a team of agents as a set of decentralized finite-state-automata (FSA) policy controllers, $\mathcal{C}^s = \{\mathcal{C}^s_k\}_{k=1}^n$, each of which characterizes a single agent's corresponding part of the stratagem, $\mathcal{C}^s_k$. Each individual controller $\mathcal{C}^s_k$ has $p$ nodes $\{q_k^i\}_{i=1}^p$ and there are two probabilistic functions associated with each node $q_k^i$: (a) an output function $\lambda(m_k^j | q_k^i) \triangleq \lambda^k_{ij}$ which decides the probability $\lambda^k_{ij}$ that macro-action $m_k^j \in \mathbb{M}_k$ is selected by agent $k$; and (b) a transition function $\delta(q_k^t | q_k^i, m_k^j) \triangleq \delta_{ij}^k(t)$ which determines the probability to transit from $q_k^i$ to $q_k^t$ following the execution of the selected macro-action $m_k^j$. The weights $\mathbf{w} \triangleq \{\{\lambda_{ij}^k\},\{\delta_{ij}^k(t)\}\}$ can then be optimized via simulation using the graph-based direct cross-entropy (G-DICE) optimization method described in \cite{Shayegan16} (see Figure~\ref{fig:1}).

In essence, G-DICE iteratively samples $\mathbf{w}$ from a distribution $q(\mathbf{w};\theta)$ parameterized by $\theta$ and simulates the induced policy (with respect to $\mathbf{w}$) with the opponent's tactic $\mathcal{E}^s \triangleq \{\mathcal{E}^s_k\}_{k=1}^n$ using its black-box simulator to acquire a performance estimate $\mathbb{L}(\mathcal{C}^s(\mathbf{w}),\mathcal{E}^s)$. At each iteration, a subset of samples with top performance estimates is used to update $\theta$ via maximum likelihood estimation (MLE). This process has been demonstrated empirically in \cite{Rubinstein04} to converge towards a uniform distribution over optimal values of $\mathbf{w}$. In practice, this optimization paradigm is very well-fitted to multi-robot planning scenarios since it allows us to bypass the explicit probabilistic modeling of opponent's tactic which is usually fraught with the curses of dimensionality and histories, especially in complex problem domains with large number of agents, vast action and observation spaces \cite{Shayegan16}. This method will also serve as the building block for our stratagem fusion scheme detailed in Section~\ref{fusion} below.

\section{STRATEGEM FUSION}
\label{fusion}
\begin{figure}[t]
\begin{tabular}{cc}
\hspace{-2.1mm}\includegraphics[width=3cm,height=3cm]{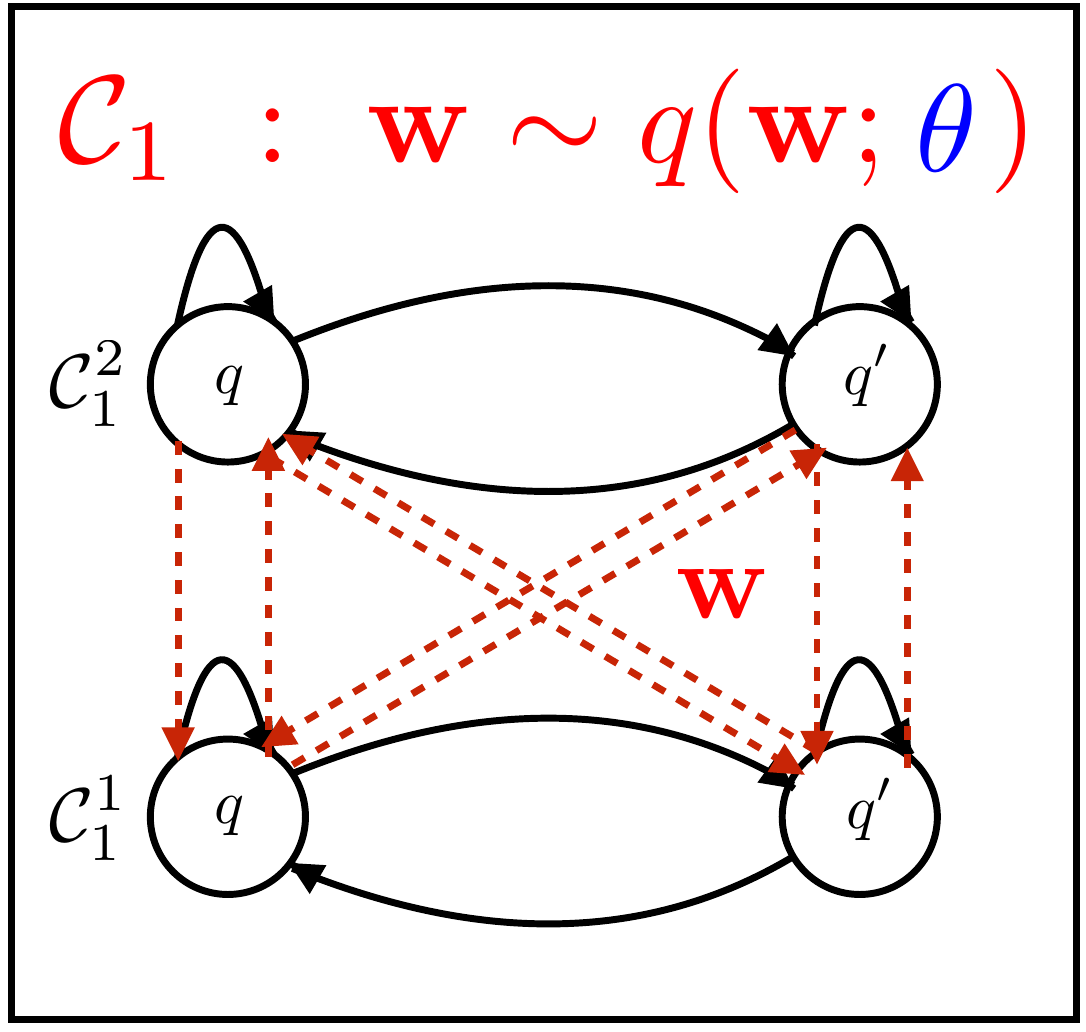} & \hspace{2mm}\includegraphics[width=5cm,height=3.025cm]{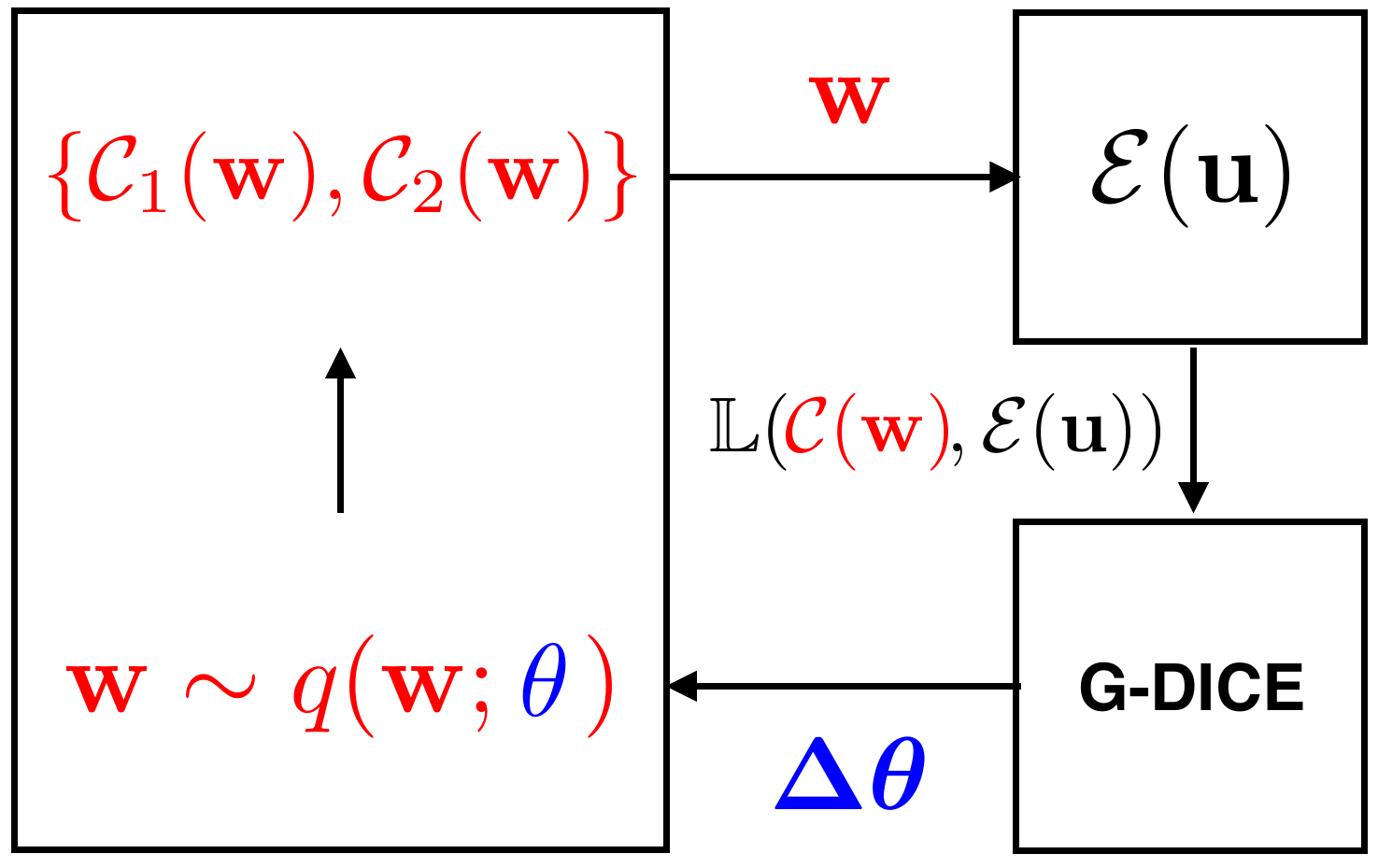}\\
(a) & (b)
\end{tabular}
\caption{(a) An agent's unified controller $\mathcal{C}_1(\mathbf{w})$ that connects its low-level controllers (i.e., stratagems) $\{\mathcal{C}_1^1,\mathcal{C}_1^2\}$ via inter-controller transitions (i.e., denote by the red, dash lines) whose weights $\mathbf{w}$ are to be optimized; and (b) a team of two agents optimizes their high-level joint controller $\{\mathcal{C}_1(\mathbf{w}),\mathcal{C}_2(\mathbf{w})\}$ via interaction with the black-box simulator of the adversaries' switching strategy $\mathcal{E}(\mathbf{u})$ given the switching weights $\mathbf{u}$.}\vspace{-3mm}
\label{fig:2}
\end{figure}

This section introduces the stratagem fusion scheme that integrates all basic stratagems (see Section~\ref{skill}) into a set of unified policies for a team of agents to collaborate effectively against the adversaries' high-level switching policies that switch opportunistically among a set of basic tactics. 
The task of stratagem fusion is then to formulate a high-level policy that can automatically detect situation changes and choose which response to follow at any point of decision to adapt effectively to new situations (e.g., the adversaries decide to switch to a different tactic) and consequently, maximize its expected performance. To achieve this, we model the team's high-level policy as a set of unified controllers, each of which characterizes a single agent's high-level individual policy that results from connecting its low-level controllers via inter-controller transitions (see Figure~\ref{fig:2}). This essentially allows the agents to change their strategic choices during real-time execution by transiting between different nodes of different controllers. The weights associated with these transitions therefore regulate the switching decision of the high-level controller and need to be optimized. If we know exactly how the adversaries change their tactics (i.e., their black-box simulators) in response to our strategic choices, these weights can be optimized using the same approach described in Section~\ref{skill} (see Figure~\ref{fig:2}b). 

In practice, however, the adversaries' switching mechanism is often unknown or highly non-trivial to characterize, especially in decentralized settings where their strategic choices are largely influenced by their limited observation and communication capacities, which are also unknown. Existing works \cite{NghiaIJCAI13a,Doshi2005,Doshi2008,Doshi2010} that attempt to reason explicitly about the adversaries' strategic rationalities are therefore impractical and less robust in situations where irrational choices arise due to limited cognitive abilities and lack of communication. This motivates us to consider a more reasonable approach to formulate a robust policy that works well on average when tested against all possible high-level strategies of the adversaries. To achieve this, the adversaries' switching policies are similarly modeled as high-level controllers that connect low-level controllers representing their basic tactics using inter-controller transitions as illustrated in Figure~\ref{fig:2}a. The weights of these inter-controller transitions (that regulate switching decisions) are then treated as random variables distributed by a known distribution. Thus, instead of optimizing our agents' switching weights with respect to a single realization of the adversaries' inter-controller transitions, we optimize them with respect to the distribution of these switching weights to embrace their uncertainty.\\ 

\noindent Formally, let $\mathcal{C} = \{\mathcal{C}_k\}_{k=1}^n$ and $\mathcal{E} = \{\mathcal{E}_k\}_{k=1}^n$ denote the sets of high-level controllers for the teams of collaborative agents and adversaries, respectively, where $\mathcal{C}_k = \{\mathcal{C}_k^s\}_s$ ($\mathcal{E}_k = \{\mathcal{E}_k^s\}_s$) denotes a single agent's (adversary's) individual switching policy. Let $\mathbf{w} = \{\mathbf{w}_k\}_{k=1}^n$ and $\pi(\mathbf{u})$ denote the weights associated with inter-controller transitions of $\mathcal{C} = \{\mathcal{C}_k\}_{k=1}^n$ and the distribution over random weights $\mathbf{u} = \{\mathbf{u}_k\}_{k=1}^n$ that regulates the switching decision of $\mathcal{E} = \{\mathcal{E}_k\}_{k=1}^n$, respectively. Our approach proposes to optimize $\mathbf{w}$ such that the expected performance of the induced high-level controller $\mathcal{C}(\mathbf{w})$ when tested against a random adversary $\mathcal{E}(\mathbf{u})$ distributed by $\pi(\mathbf{u})$ is maximized:
\begin{eqnarray}
\hspace{-5.5mm}\mathbf{w}^\ast \hspace{-2mm}&=&\hspace{-2mm} \argmax_\mathbf{w}\ \left(\mathbb{L}(\mathbf{w}) \ \triangleq\ \mathbb{E}_{\mathbf{u} \sim \pi(\mathbf{u})}\Big[\mathbb{L}\left(\mathcal{C}(\mathbf{w}), \mathcal{E}(\mathbf{u})\right)\Big] \right)\ ,\label{eq:3}
\end{eqnarray}
where $\mathbb{L}(\mathcal{C}(\mathbf{w}),\mathcal{E}(\mathbf{u}))$ denotes the simulated performance of $\mathcal{C}(\mathbf{w})$ against $\mathcal{E}(\mathbf{u})$. However, since we can only access the value of $\mathbb{L}(\mathcal{C}(\mathbf{w}),\mathcal{E}(\mathbf{u}))$ via simulation, solving \eqref{eq:3} requires simulating $\mathcal{C}(\mathbf{w})$ against infinitely many candidates of $\mathcal{E}(\mathbf{u})$ and is therefore intractable. To sidestep this intractability, we instead exploit the following surrogate objective function,
\begin{eqnarray}
\hspace{-7.5mm}\widehat{\mathbf{w}} \hspace{-2mm}&=&\hspace{-2mm} \argmax_\mathbf{w}\ \left(\widehat{\mathbb{L}}(\mathbf{w}) \ \triangleq\ \frac{1}{m}\sum_{i=1}^m\mathbb{L}\left(\mathcal{C}(\mathbf{w}), \mathcal{E}(\mathbf{u}^{(i)})\right) \right)\ ,\label{eq:4}
\end{eqnarray}
where $\{\mathbf{u}^{(i)}\}_{i=1}^m$ are i.i.d samples drawn from $\pi(\mathbf{u})$. Intuitively, these are the potential candidates for the adversaries' switching weights that can be identified in advance using the domain expert's knowledge. We can now solve \eqref{eq:4} using G-DICE \cite{Shayegan16} (see Section~\ref{skill}) with a meta black-box that aggregates the feedback of each black-box $\mathcal{E}(\mathbf{u}^{(i)})$.

\section{THEORETICAL ANALYSIS}
\label{guarantee}
This section derives performance guarantees for the above stratagem fusion scheme (Section~\ref{fusion}) which depend on the solution quality of the graph-based direct cross-entropy (G-DICE) optimization method described in \cite{Shayegan16}. 
To enable the analysis, we put forward the following assumption:\\

\noindent {\bf Assumption 1.} Let $\widehat{\mathbb{L}}(\mathbf{w})$ denote an arbitrary black-box function being optimized via simulation with G-DICE using \eqref{eq:4}. Let $\mathbb{U} \triangleq \{\mathbf{w}\ |\ \widehat{\mathbb{L}}(\mathbf{w}) = \max_{\mathbf{w}'} \widehat{\mathbb{L}}(\mathbf{w}')\}$ denotes the set of optimal solutions to \eqref{eq:4}. Then, let $p(\mathbf{w})$ and $q(\mathbf{w};\theta)$ denote the uniform distribution over $\mathbb{U}$ and the sampling distribution of G-DICE parameterized by $\theta$ (see Section~\ref{skill}). For any $\delta \in (0, 1)$, there exists a non-decreasing sequence $\{\epsilon_{n,\delta}\}_{n=1}^{\infty}$ for which:
\begin{eqnarray}
\mathrm{Pr}\left(\mathbb{D}_{\mathrm{KL}}\Big(q\left(\mathbf{w};\theta^\ast\right)\| p(\mathbf{w})\Big)\ \leq\ \epsilon_{n,\delta}\right) &\geq& 1 - \delta \ ,\nonumber
\end{eqnarray}
where $n$ and $\theta^\ast$ denote the size of $\mathbf{w}$ and the optimal parameterization of $q(\mathbf{w}; \theta)$ found by G-DICE, respectively.\\ 

This is a reasonable assumption to make since it has been previously demonstrated that the underlying cross-entropy optimization process of G-DICE empirically causes $q(\mathbf{w}; \theta)$ to converge towards the uniform distribution $p(\mathbf{w})$ over optimal values of $\mathbf{w}$ \cite{Shayegan16,Rubinstein04}. Then, let $\mathbb{L}(q) \triangleq \mathbb{E}_\mathbf{w}[\mathbb{L}(\mathbf{w})]$ (with $\mathbb{L}(\mathbf{w})$ defined in \eqref{eq:3}) denote the expected performance of $\mathcal{C}(\mathbf{w})$ when $\mathbf{w}$ is drawn randomly from $q(\mathbf{w}; \theta^\ast)$, we are interested in the gap between $\mathbb{L}(q)$ and $\mathbb{L}(\mathbf{w}^\ast)$ (see Eq~\eqref{eq:3}), the latter of which is the best performance that can be achieved. Thus, this gap essentially characterizes the near-optimality of $q(\mathbf{w}; \theta^\ast)$, which are bounded below. To do this, we first establish the following results in Lemmas 1 and 2 that bound the difference between the generalized performance of $q$ (i.e., $\mathbb{L}(q)$) and its empirical version (i.e., the average performance $\widehat{\mathbb{L}}(q)$ when tested against a finite set of adversary candidates). Lemma 3 is then established to bridge the gap between the $\widehat{\mathbb{L}}(q)$ and $\mathbb{L}(\mathbf{w}^\ast)$. The main result that bounds the performance gap between $\mathbb{L}(q)$ and $\mathbb{L}(\mathbf{w}^\ast)$ is then derived in Theorem 1 as a direct consequence of the previous Lemmas.\\

{\noindent\bf Lemma 1.} For any sampling distribution $q(\mathbf{w};\theta)$, let $\widehat{\mathbb{L}}(q) \triangleq \mathbb{E}_\mathbf{w}\left[\widehat{\mathbb{L}}(\mathbf{w})\right]$, with $\widehat{\mathbb{L}}(\mathbf{w})$ defined in \eqref{eq:4}, denotes the empirical performance of $\mathcal{C}(\mathbf{w})$ where $\mathbf{w}$ is randomly drawn from $q(\mathbf{w}; \theta)$. Then, it follows that with probability at least $1 - \delta$ over the choice of candidates $\{\mathbf{u}^{(i)}\}_{i=1}^m$ for the adversaries' switching weights,
\begin{eqnarray}
\hspace{-8.5mm}\mathbb{L}(q) &\leq& \widehat{\mathbb{L}}(q) + \left(\frac{\mathbb{D}_{\mathrm{KL}}\left(q(\mathbf{w};\theta) \|p(\mathbf{w})\right) + \log \frac{4m}{\delta}}{2m - 1}\right)^{\frac{1}{2}} \label{eq:5}
\end{eqnarray}
holds universally for all possible $q(\mathbf{w}; \theta)$ where $p(\mathbf{w})$ denotes the uniform distribution over the set $\mathbb{U}$ of optimal choice of $\mathbf{w}$ for \eqref{eq:4}, i.e., $\mathbb{U} \triangleq \{\mathbf{w} \ |\ \widehat{\mathbb{L}}(\mathbf{w}) = \max_{\mathbf{w}'} \widehat{\mathbb{L}}(\mathbf{w}')\}$. \\


\noindent Exploiting the result of Lemma 1, we can further derive a tighter and domain specific bound on the difference between the generalized and empirical performance of our stratagem fusion scheme (see Section~\ref{fusion}) that incorporates the empirical optimality of G-DICE (see Assumption 1):\\

{\noindent\bf Lemma 2.} Let $q \ \triangleq\ q(\mathbf{w};\theta^\ast)$ denotes the optimal sampling distribution found by G-DICE \cite{Shayegan16}. Let $r$ denotes the number of stratagems of each agent (Section~\ref{skill}) and let $k$ denotes the number of nodes in each agent's individual specialized controller $\mathcal{C}_k^s$. It then follows that with probability at least $1 - \delta$,\vspace{-3mm}
\begin{eqnarray}
\hspace{-8mm}\mathbb{L}(q) &\leq& \widehat{\mathbb{L}}(q) \ +\ \left(\frac{\epsilon_{h,\frac{\delta}{2}} + \log \frac{8m}{\delta}}{2m - 1}\right)^{\frac{1}{2}} \ , \label{eq:6}\vspace{-3mm}
\end{eqnarray}
where $\mathbb{L}(q)$ and $\widehat{\mathbb{L}}(q)$ are defined in Lemma 1, $h = O\left(nr(r-1)k^2\right)$ and $\delta \in (0, 1)$.\\


\noindent Lemmas 1 and 2 thus bound the performance gap between $\mathbb{L}(q)$ and $\widehat{\mathbb{L}}(q)$. To relate $\mathbb{L}(q)$ to $\mathbb{L}(\mathbf{w}^\ast)$, we need to bound the gap between $\widehat{\mathbb{L}}(q)$ and $\mathbb{L}(\mathbf{w}^\ast)$, which is detailed below. \\

{\noindent\bf Lemma 3.} Let $q \ \triangleq\ q(\mathbf{w};\theta^\ast)$ denotes the optimal sampling distribution found by G-DICE \cite{Shayegan16} and $\delta \in (0, 1)$, then with probability at least $1 - \delta$,\vspace{-3mm}
\begin{eqnarray}
\widehat{\mathbb{L}}(q) &\leq& \mathbb{L}(\mathbf{w}^\ast) \ +\ \left(\frac{\log\frac{1}{\delta}}{2m}\right)^{\frac{1}{2}} \ , \label{eq:9}\vspace{-3mm}
\end{eqnarray}
where $\widehat{\mathbb{L}}(q)$ is defined in Lemma 1.\\


\noindent Using these results, the key result can be stated and proven:\\

{\noindent\bf Theorem 1.} Let $q \ \triangleq\ q(\mathbf{w};\theta^\ast)$ denotes the optimal sampling distribution found by G-DICE \cite{Shayegan16} and $\mathbf{w}^\ast$ denotes the optimal solution to \eqref{eq:3}. $\mathbb{L}(\mathbf{w}^\ast)$ thus represents the best possible performance and with probability at least $1 - \delta$,\vspace{-2mm}
\begin{eqnarray}
\hspace{-8mm}\mathbb{L}(q) &\leq& \mathbb{L}(\mathbf{w}^\ast) \ +\ 2\left(\frac{\epsilon_{h,\frac{\delta}{4}} + \log \frac{16m}{\delta}}{2m - 1}\right)^{\frac{1}{2}} \ , \label{eq:12}\vspace{-2mm}
\end{eqnarray}
where $\mathbb{L}(q)$ is defined in Lemma 1, $h = O\left(nr(r-1)k^2\right)$. Due to the limited space, all proofs of the above results are deferred to the appendix of the extended version of this paper at \url{https://www.dropbox.com/s/ao7onnpq52t3ar3/icra18.pdf?dl=0}.



\section{EXPERIMENTS}
\label{experiment}
This section presents an adversarial, multi-robot Capture-The-Flag (CTF) domain adapted from its original domain in \cite{Murray03} to demonstrate the effectiveness of our stratagem fusion framework in Section~\ref{fusion}. The specific domain setup for our CTF variant is detailed in Section~\ref{CTF} below. 

\begin{figure*}[ht]
\centering
\begin{tabular}{cc}
\includegraphics[width=8.5cm,height=4.8cm]{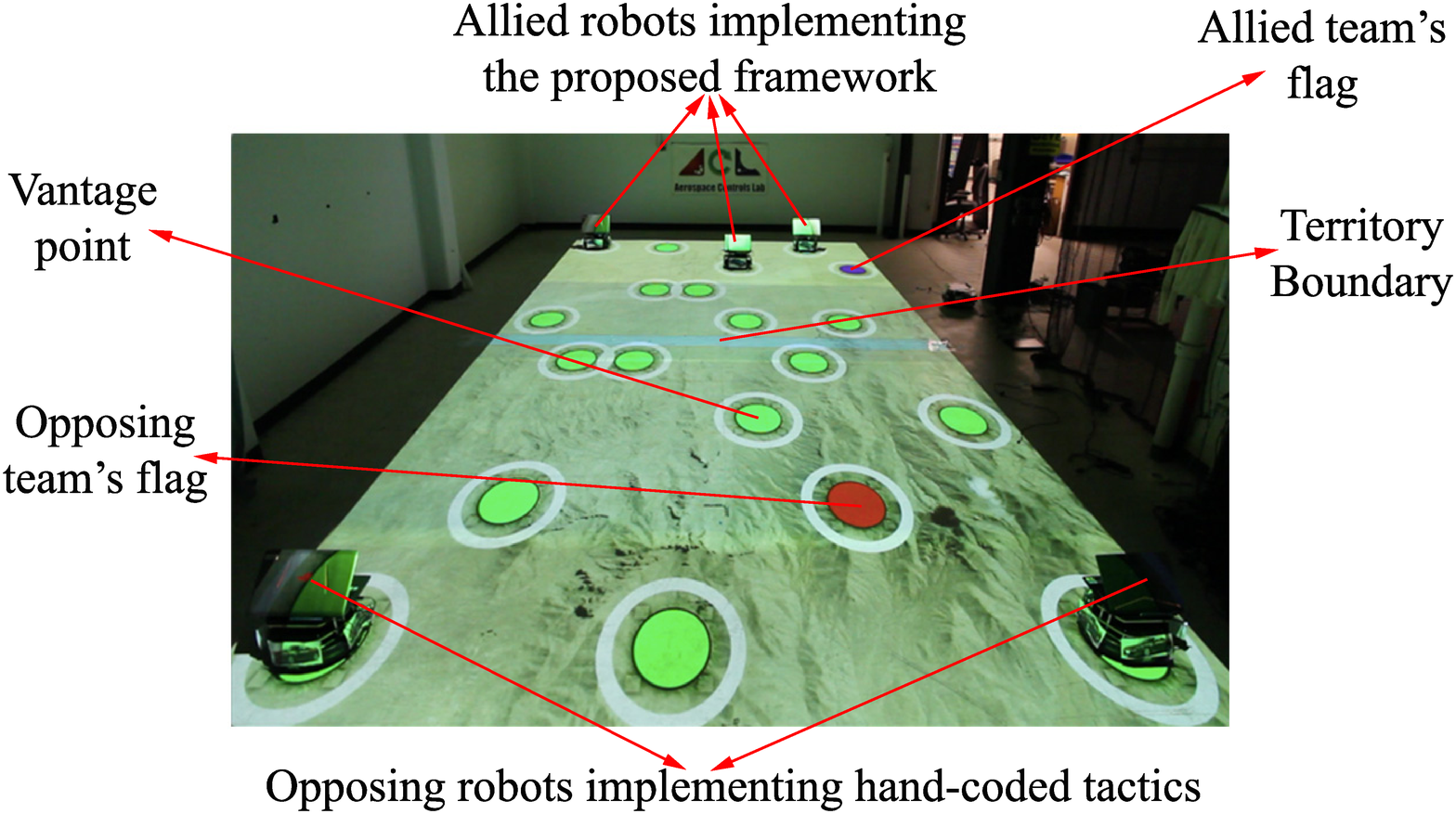} & \includegraphics[width=8.5cm,height=4.65cm]{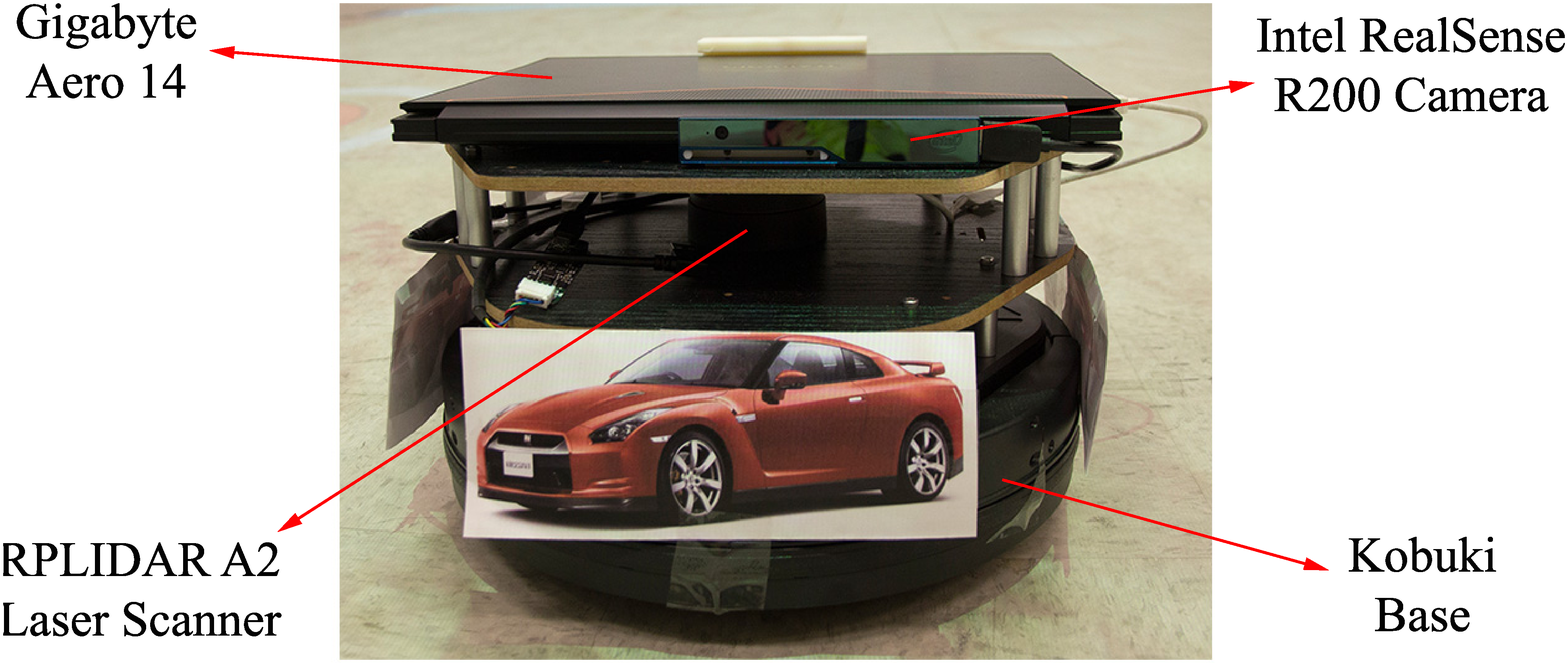} \vspace{-2mm}\\
(a) & (b)\vspace{-2mm}
\end{tabular}
\caption{Figures of (a) Capture-The-Flag (CTF) domain setup; and (b) hardware configuration of the experimented robots: each robot is built from the Kobuki base of the TurtleBot $2$ with on-board processing unit (Gigabyte Aero 14 laptop with Intel Core i$7$-$7700$HQ quad-core CPU and NVIDIA GTX $1060$ GPU with $6$GB RAM) as well as sensory devices including (1) Intel RealSense Camera (R$200$) Developer Kit ($130$mm x $20$mm x $7$mm) with Depth/IR: Up to $640 \times 480$ resolution at $60$ FPS $\&$ RGB: $1080$p at $30$ FPS; and (2) Omnidirectional RPLIDAR A2 with $4000$ samples/s (10Hz) and 8/16m range.}\vspace{-5mm}
\label{fig:5}
\end{figure*}

\subsection{Capture-The-Flag Domain}
\label{CTF}
The domain settings for Capture-The-Flag are shown in Fig.~\ref{fig:5}a, which depicts a competitive scenario between two teams of decentralized, collaborative robots. Each team has $2-3$ robots and the two teams divide the environment into two parts separated by a horizontal boundary (the cyan line in Fig.~\ref{fig:5}a), each of which belongs to one team (e.g., red $\&$ blue). There are $10$ vantage points within each team's territory. One of which contains the flag of the team (e.g., the red and blue circles in Fig~.\ref{fig:5}a denote the locations of the flags for the red and blue team, respectively). Each team, however, only knows the location of its own flag, thus making observations necessary to correctly detect the enemy's flag. The rule of the game is for each team to defend its own flag while seeking to capture the flag of the opposing team without getting caught. The game ends when one team successfully captures the flag of the opposing team. To achieve this, each team of agents need to coordinate their movements between vantage points to reach the opposing team's flag and at the same time, avoid being seen by opposing agents. If an agent engages an opposing agent on foreign territory, its team will be charged with a penalty. The particular macro-actions and -observations available for each robot are detailed below, which feature a wide range of interesting observations and patterns of collaborative attack and defend for the opposing robots:\vspace{-2mm}\\


{\noindent \bf Macro-Actions.} There are $4$ classes of macro-actions available to each robot at any decision time: (a) $\mathbf{Move}(p)$ which invokes a collision avoidance navigation procedure that directs the robot to vantage point $p$ from its current location; (b) $\mathbf{Sentry}(p_1, p_2, p_3)$ which directs the robot to vantage point $p_1$ and then lets it stay in a closed-loop moving from $p_1$ to $p_2$ to $p_3$ and back to $p_1$. There are $5$ predefined instances for each team; (c) $\mathbf{Pincer}(p_1, p_2, p_3, p_4)$ which directs the robot to vantage point $p_i$ (with $i$ being its role index in the team) and then $p_4$. This creates an effective pincer attack when $2$ or $3$ robots choose the same $\mathbf{Pincer}$ instance. There are $3$ different $\mathbf{Pincer}$ macros predefined for each team; and (d) $\mathbf{Tag}$ which allows a robot to catch an opposing robot on its own territory provided the opposing robot is within a predefined tagging range.\vspace{-2mm} \\


{\noindent \bf Macro-Observations.} There are in total $128$ macro-observations for each robot, which are generated by first collecting raw observations the environment using the robot's on-board visual recognition/detection modules and then summarizing the raw information into a $6$-dimensional observation vector. Each observation is represented as a $6$-dimensional binary vector whose components correspond to yes/no ($1/0$) answers to the following questions: (a) Is the robot residing in its own or the opposition territory? (b) does the enemy flag appear in sight? (c) is there an opposition robot in close proximity? (d) is there an opposition robot further away? (e) is there an allied robot in the vicinity? and (f) is there an observed pincer signal emitted from allied robots? The answers to these questions can be generated from the raw visual processing unit on-board each robot.\vspace{-2mm}\\

{\noindent \bf Rewards.} Finally, in order to encourage each team to discover and capture the opposition's flag as soon as possible while avoid getting tagged, a reward mechanism is implemented which issues (a) a negative reward of $-1$ to each robot at each time step; (b) a positive reward of $10$ to a team if one of its member successfully tags an enemy; (c) a negative reward of $-10$ for the entire team if one of the team member gets caught; and (d) a large award of $500$ is issued to a team when it successfully captures the opposition's flag. Conversely, this implies a large penalty of $-500$ issued to the other team who loses the flag.\vspace{-2mm}\\ 

{\noindent \bf Black-box Simulators.} In addition to the domain specification above, the allied robots also have access to a set of black-box simulators of the opposition's fundamental tactics upon which more advanced strategies might be built. In our experiments, these are constructed as tuples of individual hand-coded tactics (see Table~\ref{table:1} below) that include: (a) $\mathbf{DL}$ and $\mathbf{DR}$ which script the robot to play defensively on left and right flank of its territory using $\mathbf{Sentry}$ and $\mathbf{Move}$ macro-actions, respectively; (b) $\mathbf{DC}$ which scripts the robot to play defensively on the middle-front of the allied territory; (c) $\mathbf{AS}$ which leads the robot to a vantage point inside the opposition's territory to get an observation. Depending on the collected observation, the robot either moves to another vantage point or launch a pincer attack to a vantage point estimated to contain the opposition's flag; and (d) $\mathbf{AA}$ which is similar to $\mathbf{AS}$ except that it enables the robot to retreat to a safe place within the allied territory to gather extra observations if it observes that there is an opposing robot in close proximity. The team of allied robots however do not have access to these details and can only interact with them via a black-box interface that gives feedback on how well their strategies fare against the opposition's.


\begin{table}[t]
\caption{The opposition's team tactics $\mathcal{E}^s$ represented as combinations of individual's tactics $\{\mathcal{E}_k^s\}_k$ of $3$ robots $\mathbf{R1},\mathbf{R2}$ and $\mathbf{R3}$.}
\label{table:1}
\centering
\hspace{-0.5mm}\begin{tabular}{|l|l|l|l|}
\hline
\rule{0pt}{10pt} &\hspace{-1mm}{\bf R1} $\left(\{\mathcal{E}_1^s\}_{s=1}^4\right)$ & \hspace{-1mm}{\bf R2} $\left(\{\mathcal{E}_2^s\}_{s=1}^4\right)$ & \hspace{-1mm}{\bf R3} $\left(\{\mathcal{E}_3^s\}_{s=1}^4\right)$ \\
\hline
$\mathcal{E}^1 \triangleq (\mathcal{E}^1_1, \mathcal{E}^1_2, \mathcal{E}^1_3)$ &\hspace{-1mm}$\mathcal{E}^1_1\ \triangleq\ \mathbf{DL}$ & \hspace{-1mm}$\mathcal{E}^1_2\ \triangleq\ \mathbf{DC}$ & \hspace{-1mm}$\mathcal{E}^1_3\ \triangleq\ \mathbf{DR}$ \\
\hline
$\mathcal{E}^2 \triangleq (\mathcal{E}^2_1, \mathcal{E}^2_2, \mathcal{E}^2_3)$ &\hspace{-1mm}$\mathcal{E}^2_1\ \triangleq\ \mathbf{DL}$ & \hspace{-1mm}$\mathcal{E}^2_2\ \triangleq\ \mathbf{AS}$ & \hspace{-1mm}$\mathcal{E}^2_3\ \triangleq\ \mathbf{DR}$ \\
\hline
$\mathcal{E}^3 \triangleq (\mathcal{E}^3_1, \mathcal{E}^3_2, \mathcal{E}^3_3)$ &\hspace{-1mm}$\mathcal{E}^3_1\ \triangleq\ \mathbf{AA}$ & \hspace{-1mm}$\mathcal{E}^3_2\ \triangleq\ \mathbf{DC}$ & \hspace{-1mm}$\mathcal{E}^3_3\ \triangleq\ \mathbf{AS}$ \\
\hline
$\mathcal{E}^4 \triangleq (\mathcal{E}^4_1, \mathcal{E}^4_2, \mathcal{E}^4_3)$ &\hspace{-1mm}$\mathcal{E}^4_1\ \triangleq\ \mathbf{AS}$ & \hspace{-1mm}$\mathcal{E}^4_2\ \triangleq\ \mathbf{AA}$ & \hspace{-1mm}$\mathcal{E}^4_3\ \triangleq\ \mathbf{AS}$ \\
\hline
\end{tabular}\vspace{-4mm}
\end{table}

\subsection{Experiment: Generating Basic Stratagems}
\label{LSP}
\begin{figure}[t]
	\vspace{-2mm}
	\begin{tabular}{cc}
		\hspace{-4mm}\includegraphics[width=4.5cm]{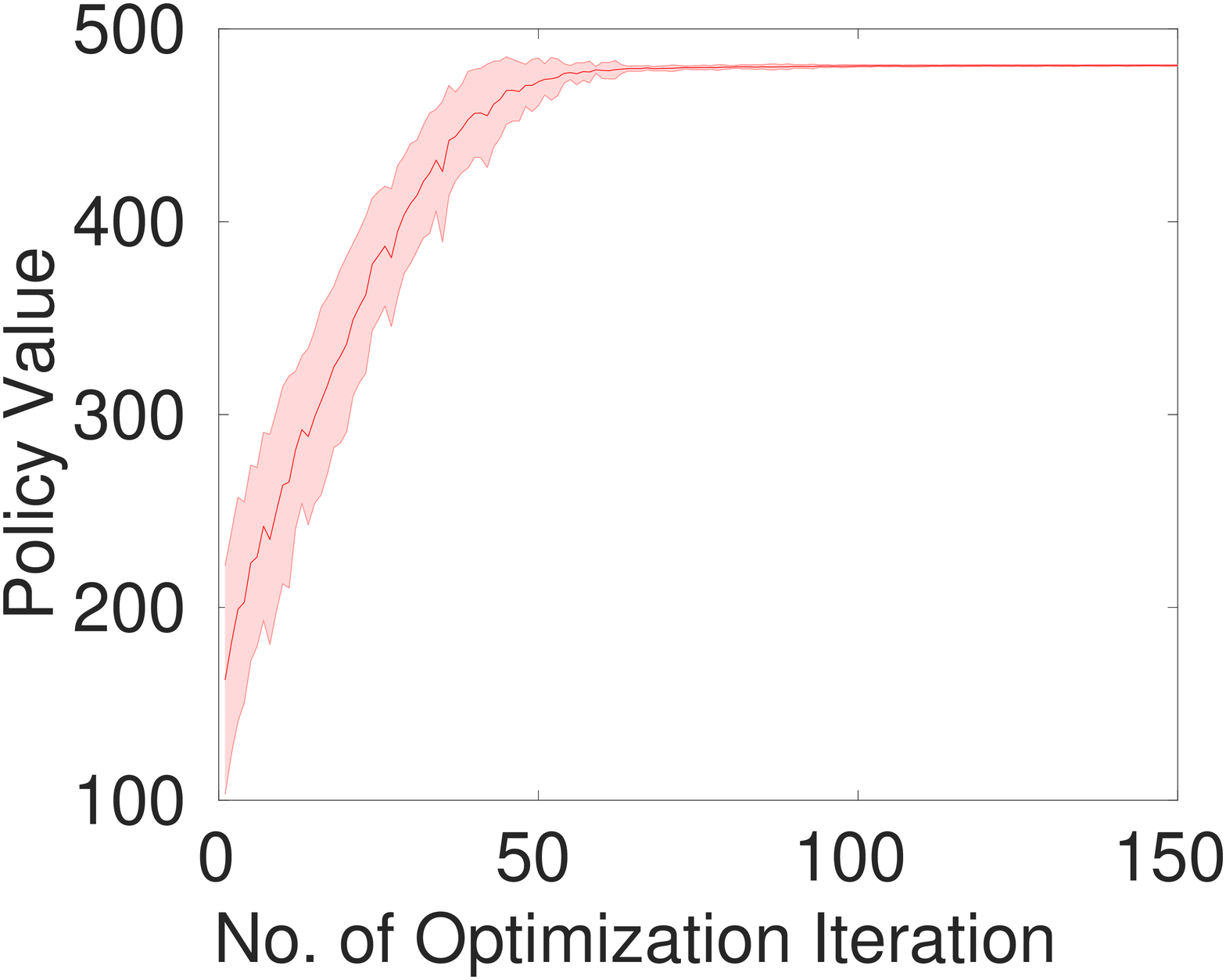} &
		\hspace{-4mm}\includegraphics[height=3.6cm,width=4.5cm]{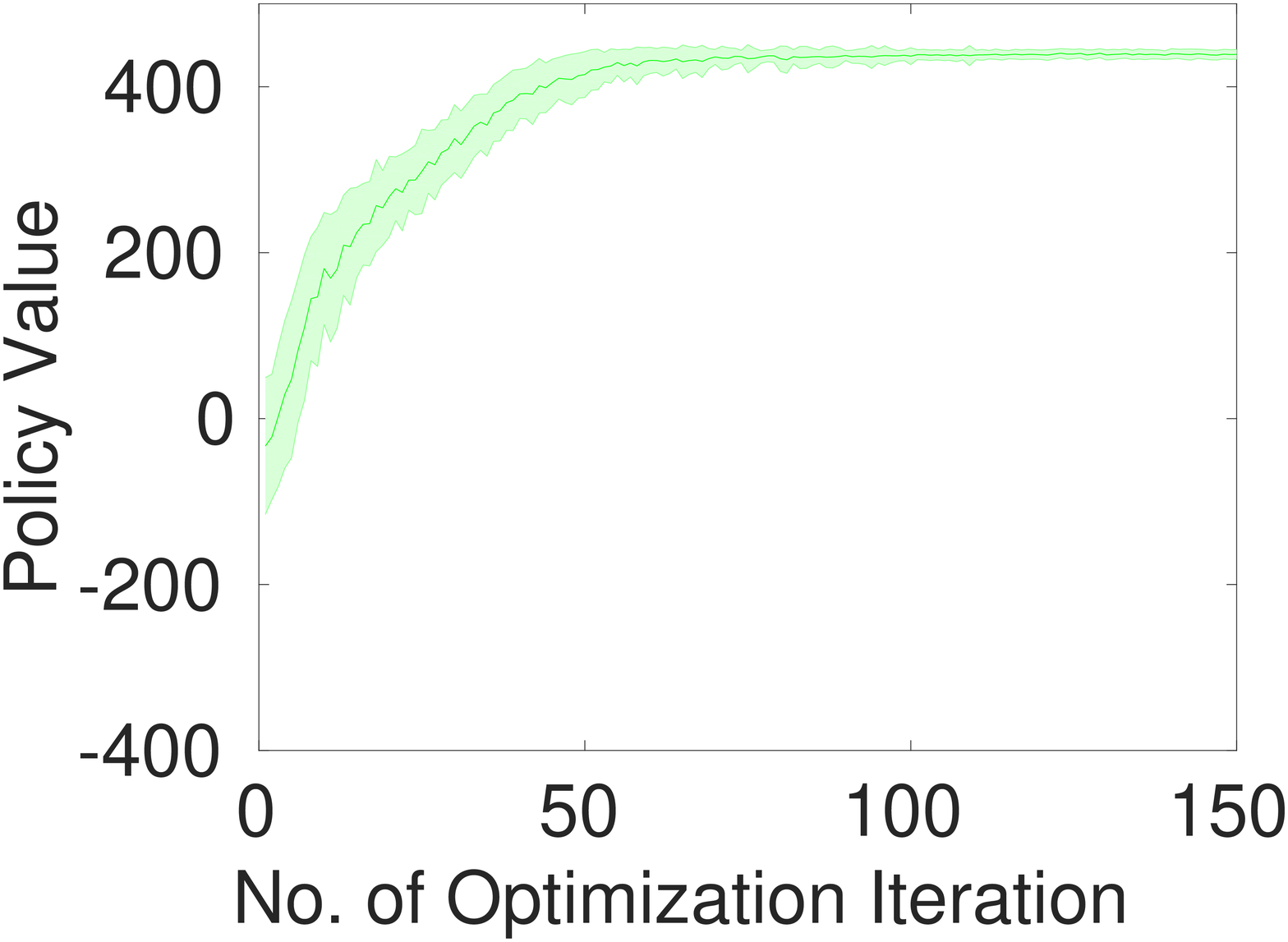} \\ \vspace{-1mm}
		\hspace{-1mm}(a) & \hspace{-1mm}(b)\vspace{-0mm} \\
		\hspace{-4mm}\includegraphics[width=4.5cm]{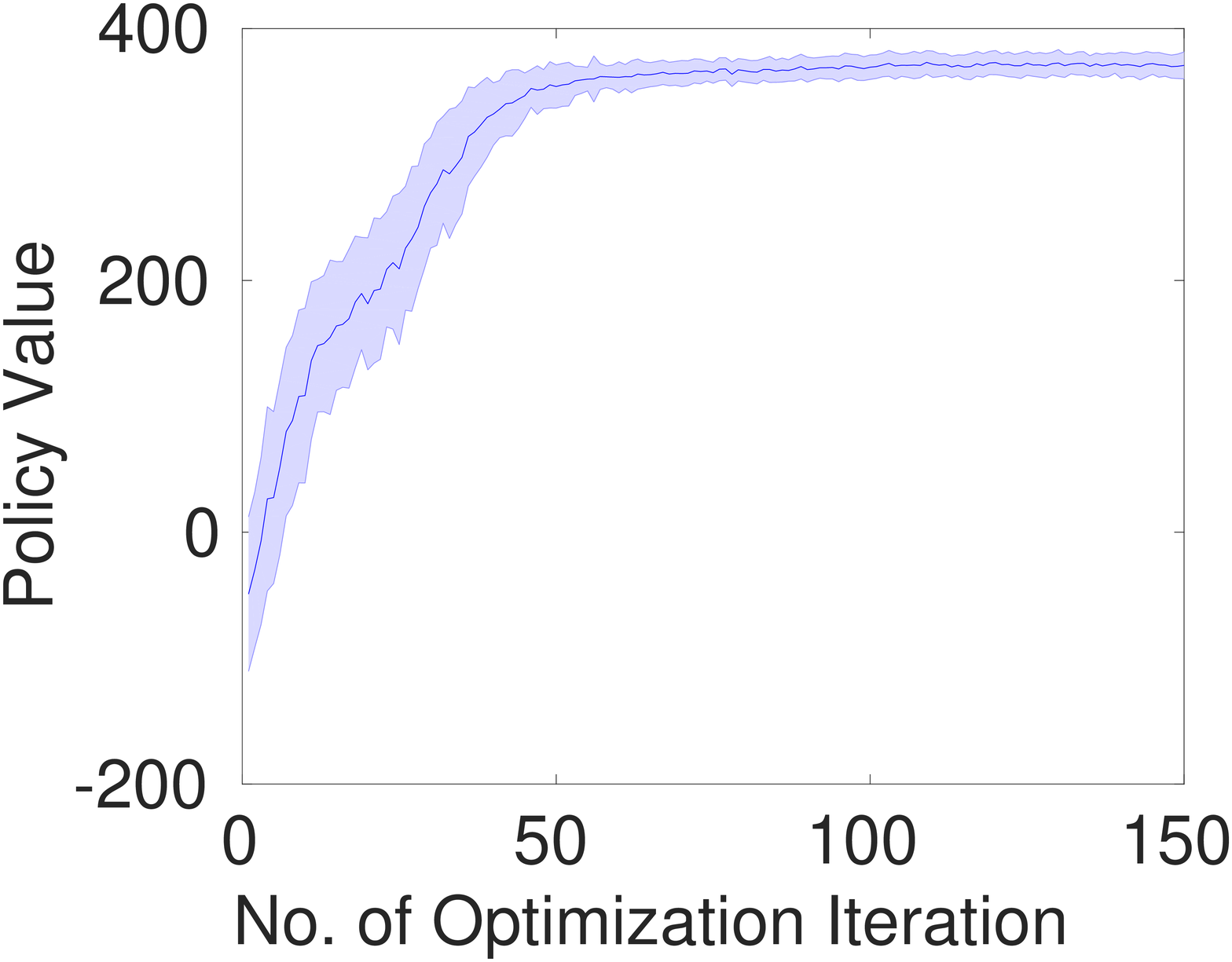} &
		\hspace{-4mm}\includegraphics[width=4.5cm]{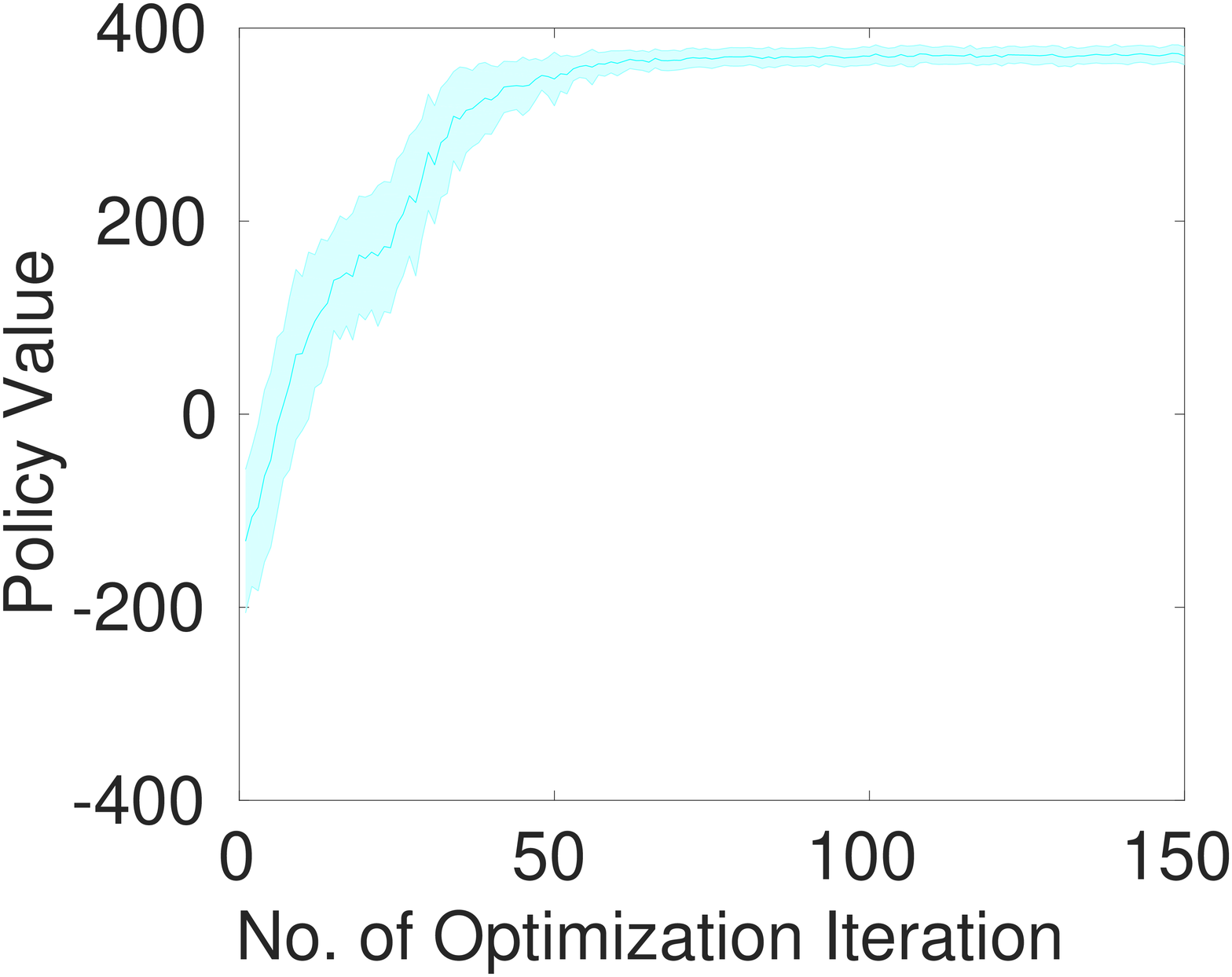} \\ \vspace{-1mm}
		\hspace{-1mm}(c) & \hspace{-1mm}(d) \\\vspace{-4mm}
	\end{tabular}
	\caption{Graphs of each stratagem's increasing performance quality in the no. of optimization iterations when optimized against the opposition's corresponding tactic of (a) $(\mathbf{DL},\mathbf{DC},\mathbf{DR})$, (b) $(\mathbf{DL},\mathbf{AS},\mathbf{DR})$, (c) $(\mathbf{AA},\mathbf{DC},\mathbf{AS})$ and (d) $(\mathbf{AS}, \mathbf{AA}, \mathbf{AS})$ (see Table~\ref{table:1}). The shaded area represents the confidence interval of the average performance.}\vspace{-6mm}
	\label{fig:3}
\end{figure}

\noindent To learn the fundamental stratagems to counter the opposition's basic tactics as described in Section~\ref{CTF}, we construct separate MacDec-POMDPs (see Section~\ref{DecSPOMDP}) that encapsulates the opposition's corresponding tactic simulator $\mathcal{E}^s \triangleq \{\mathcal{E}^s_k\}_k$. The corresponding stratagem can then be formulated and computed as decentralized FSA controllers $\mathcal{C}^s \triangleq \{\mathcal{C}^s_k\}_k$ (Section~\ref{skill}) that optimizes these MacDec-POMDPs. This is achieved via a recently developed graph-based direct cross-entropy (G-DICE) stochastic optimization method of \cite{Shayegan16}. Fig~\ref{fig:3} shows that the empirical performance of each stratagem $\mathcal{C}^s$ when tested against the corresponding opposition's tactic $\mathcal{E}^s$ increases and converges rapidly to the optimal performance when we increase the number of optimization iterations. Table~\ref{table:2} then reports the averaged performance (with standard errors) of each stratagem when tested against all other opposition's tactics over $1000$ independent simulations. The results interestingly show that the quality of each stratagem $\mathcal{C}^s$ decreases significantly when tested against other opposition's tactics $\{\mathcal{E}^{s'}\}_{s'\ne s}$ that it was not optimized to interact with (see Table~\ref{table:2}'s first $4$ rows and columns). This implies a performance risk when applying a single stratagem against non-stationary opponent with switching tactics: The applied stratagem might no longer be optimal when the opponent switches to a new tactic. This necessitates design of agents which can detect and respond aptly when the opponents change their tactics which constitutes the main contribution of our work (Section~\ref{fusion}). Its effectiveness is demonstrated next in Section~\ref{iss}.

\begin{table*}[t]
\caption{Average performance (with standard errors) of the robots' basic stratagems $\mathcal{C}^1, \mathcal{C}^2, \mathcal{C}^3, \mathcal{C}^4$ and switching policy $\mathcal{C}(\mathbf{w})$ when tested against the opposition's basic tactics $\mathcal{E}^1, \mathcal{E}^2, \mathcal{E}^3, \mathcal{E}^4$ (see Table~\ref{table:1}) and switching tactic $\mathcal{E}(\mathbf{u})$ with switching weights $\mathbf{u}$. The switching policy $\mathcal{C}(\mathbf{w})$ is learned assuming access to a blackbox simulator of $\mathcal{E}(\mathbf{u})$ (see Section~\ref{skill}).}
\label{table:2}
\centering
\begin{tabular}{|l|M{2.90cm}|M{2.90cm}|M{2.90cm}|M{2.90cm}|M{2.90cm}|}
\hline
\rule{0pt}{10pt}  & $\mathcal{E}^1 = (\mathbf{DL},\mathbf{DC},\mathbf{DR})$ & $\mathcal{E}^2 = (\mathbf{DL},\mathbf{AS},\mathbf{DR})$ & $\mathcal{E}^3 = (\mathbf{AA},\mathbf{DC},\mathbf{AS})$ & $\mathcal{E}^4 = (\mathbf{AS}, \mathbf{AA}, \mathbf{AS})$ & $\mathcal{E}(\mathbf{u})$ \\ 
\hline
$\mathcal{C}^1$ & $\mathbf{481.235 \pm 0.119}$ & $405.737 \pm 0.933$ & $184.081 \pm 1.453$ & $126.665 \pm 1.277$ & $329.598 \pm 1.137$ \\
\hline
$\mathcal{C}^2$ & $450.004 \pm 1.217$ & $\mathbf{439.339 \pm 0.699}$ & $191.609 \pm 1.279$ & $97.129 \pm 1.392$ &  $295.037 \pm 1.302$\\
\hline
$\mathcal{C}^3$ & $296.436 \pm 2.616$ & $139.034 \pm 1.573$ & $\mathbf{374.477 \pm 1.049}$ & $190.408 \pm 1.285$ &  $263.993 \pm 1.343$\\
\hline
$\mathcal{C}^4$ & $323.481 \pm 2.463$ & $218.717 \pm 1.432$ & $352.924 \pm 0.924$ & $\mathbf{375.485 \pm 0.893}$ & $309.696 \pm 1.229$\\
\hline
$\mathcal{C}(\mathbf{w})$ & $469.007\pm 0.774$  & $399.731 \pm 0.949$ & $332.819 \pm 1.006$ & $301.353 \pm 1.095$ & $\mathbf{386.831 \pm 0.992}$\\
\hline
\end{tabular}
\end{table*}
 

\begin{table*}[t]
\caption{Average performance (with standard errors) of the allied robots' \emph{good-for-one} $\mathcal{C}(\mathbf{w})$ (optimized against a particular $\mathcal{E}(\mathbf{u})$) and \emph{good-for-all} $\mathcal{C}(\widehat{\mathbf{w}})$ (optimized against the entire distribution of $\mathbf{u}$ -- see Section~\ref{fusion}) switching policies when tested against unseen switching policies $\mathcal{E}(\mathbf{u}^{(1)}), \mathcal{E}(\mathbf{u}^{(2)}), \ldots, \mathcal{E}(\mathbf{u}^{(6)})$ of the opposition.}
\label{table:3}
\centering
\begin{tabular}{|M{2.1cm}|M{2.1cm}|M{2.1cm}|M{2.1cm}|M{2.1cm}|M{2.1cm}|M{2.1cm}|}
\hline
\rule{0pt}{10pt} & $\mathcal{E}(\mathbf{u}^{(1)})$ & $\mathcal{E}(\mathbf{u}^{(2)})$ & $\mathcal{E}(\mathbf{u}^{(3)})$ & $\mathcal{E}(\mathbf{u}^{(4)})$ & $\mathcal{E}(\mathbf{u}^{(5)})$ & $\mathcal{E}(\mathbf{u}^{(6)})$ \\ 
\hline
$\mathcal{C}(\mathbf{w})$ & $383.836 \pm 1.006$  & $385.482 \pm 0.892$ & $388.875 \pm 0.875$ & $388.361 \pm 0.876$ & $389.545 \pm 0.871$ & $389.824 \pm 0.869$\\
\hline
$\mathcal{C}(\widehat{\mathbf{w}})$ & $\mathbf{385.811 \pm 1.013}$ & $\mathbf{390.403 \pm 0.882}$ & $\mathbf{393.296 \pm 0.865}$ & $\mathbf{391.793 \pm 0.870}$ & $\mathbf{394.021 \pm 0.861}$ & $\mathbf{391.698\pm 0.873}$\\
\hline
\end{tabular}
\end{table*}

\subsection{Experiment: Stratagem Fusion}
\label{iss}
\noindent This section empirically demonstrates the effectiveness of our stratagem fusion framework (Section~\ref{fusion}) against more sophisticated and non-stationary/strategic opponents. In particular, we first evaluate the performance of the optimized stratagems in the previous experiments (Section~\ref{LSP}) against a team of opponents with switching tactic: each opponent independently switches its tactic based on a set of probability weights $\mathbf{u}$ (as previously described in Section~\ref{fusion}). The results (averaged over $1000$ independent runs) are reported in the last columns of Table~\ref{table:2}, which show significant decreases in the performance of each stratagem when tested against an opponent that keeps switching between tactics. This corroborates our observations earlier that a single stratagem is generally ineffective against opponents with unexpected behaviors. This can be remedied using our stratagem fusion scheme (see Fig.~\ref{fig:2}) to integrate all single stratagems into a unified (switching) policy which can perform effectively against the switching tactic of the opponents (assuming the switching weights $\mathbf{u}$ are known). The reported results in the last row of Table~\ref{table:2} in fact show that among all policies, the optimized switching policy performs best against the tactic-switching opponents and near-optimal against each stationary opponent: Its performance is, in most cases, only second (and very close) to the corresponding stratagem specifically designed to counter the opponent's tactic. \\

\begin{figure}[t]
	\vspace{-2mm}
	\begin{tabular}{cc}
		\hspace{-4mm}\includegraphics[height=3.6cm,width=4.5cm]{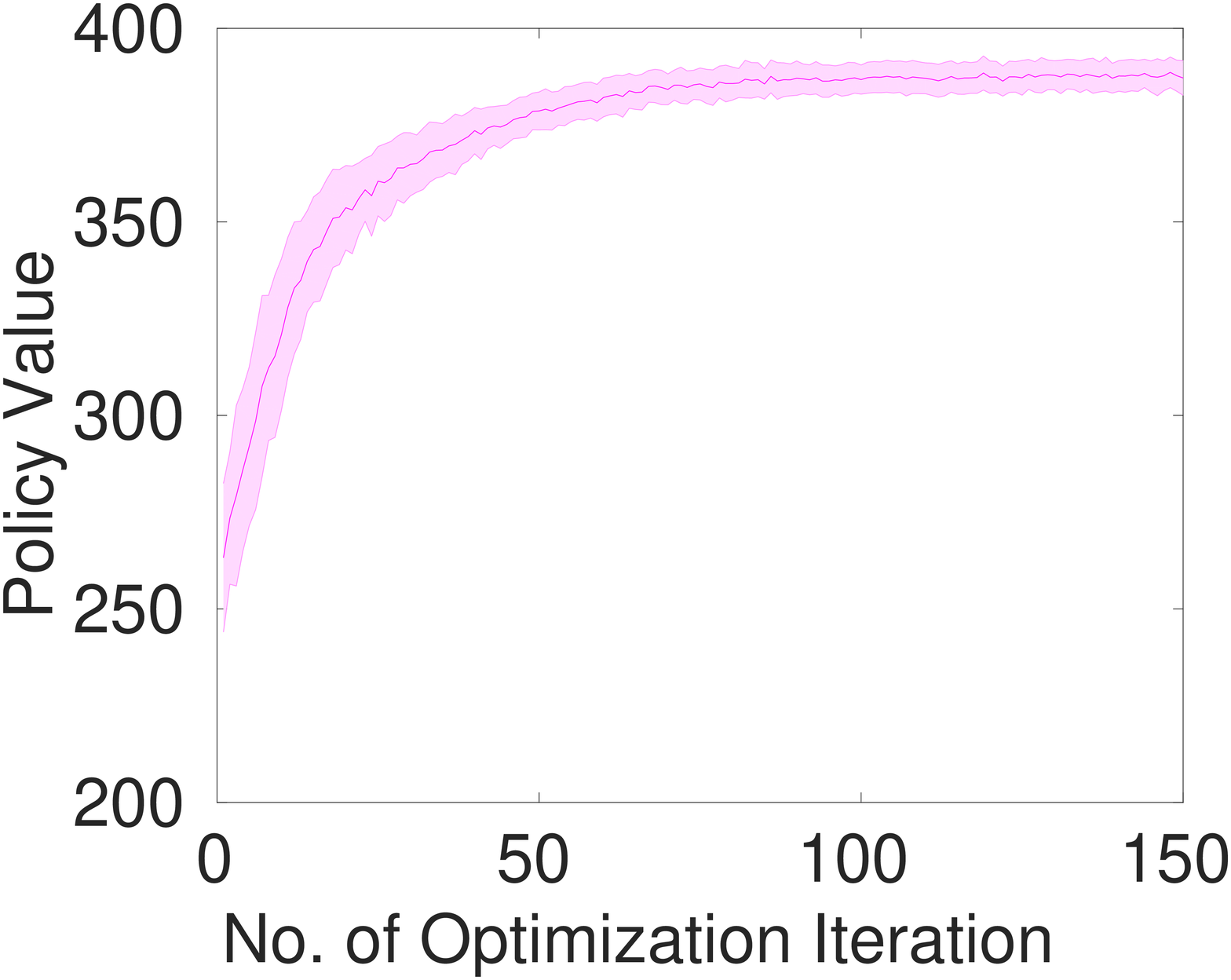} &
		\hspace{-4mm}\includegraphics[width=4.5cm]{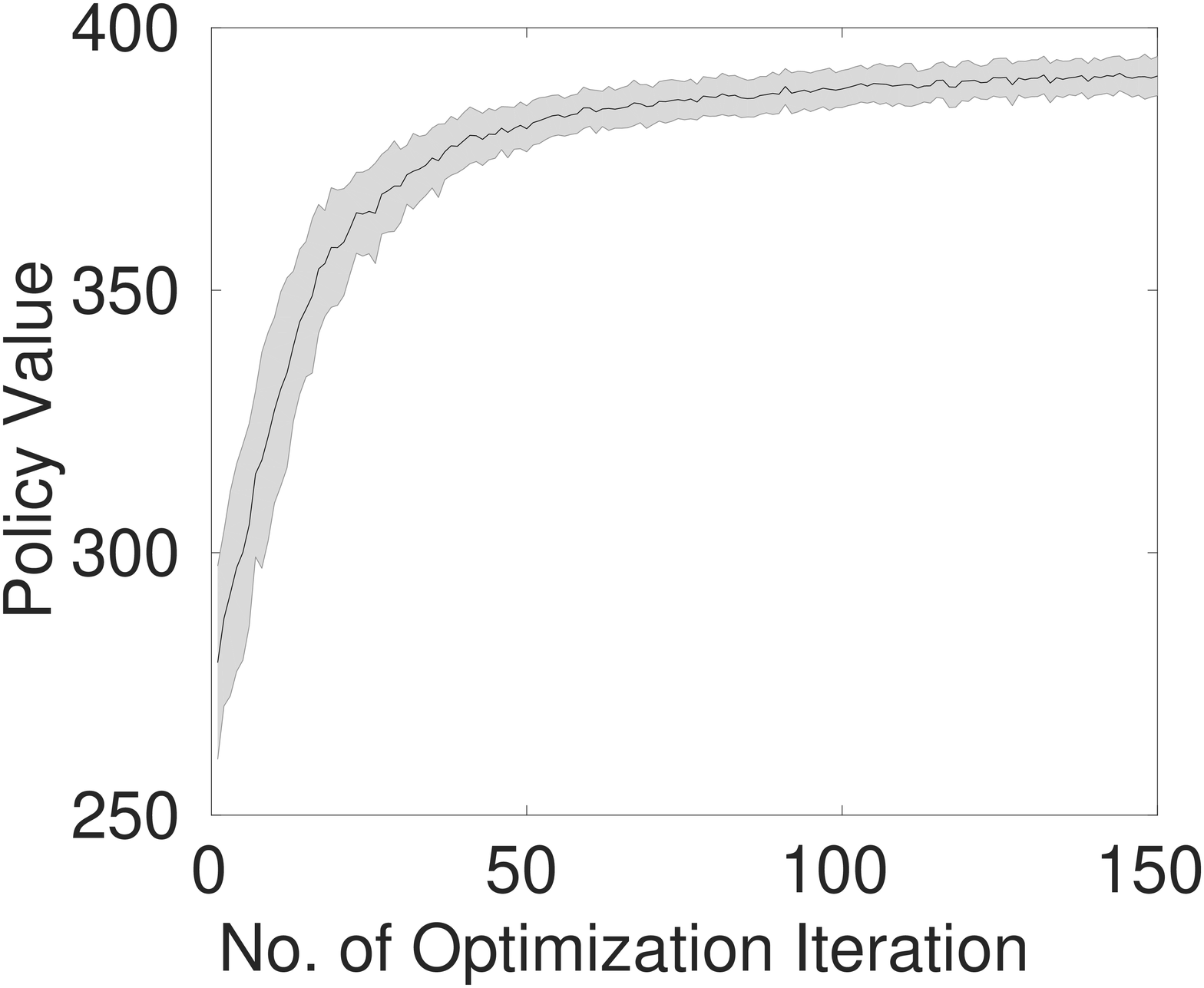} \\ \vspace{-1mm}
		\hspace{-1mm}(a) & \hspace{-1mm}(b) 
	\end{tabular}
	\caption{Graphs of the switching policy's increasing performance in the no. of optimization iterations when optimized against a switching tactic of the opposition with (a) fixed switching weights; and (b) random switching weights. The shaded area represents the confidence interval surrounding the average performance.}\vspace{-6mm}
	\label{fig:4}
\end{figure}

\noindent In practice, however, since the switching weights of the opponents are usually not known a priori, a similar problem arises when the actual weights $\mathbf{u}$ used by the opposition's switching tactic are different from those used to optimize the switching policy of the allied robots. To resolve this, our stratagem fusion scheme further treated the switching weights $\mathbf{u}$ of the opposition as random variables whose samples are either given in advance or can be drawn directly from a blackbox distribution $\pi(\mathbf{u})$. A \emph{good-for-all} switching policy $\mathcal{C}(\widehat{\mathbf{w}})$ can thus be computed using our sampling method in Section~\ref{fusion} (specifically, see Eq.~\eqref{eq:4}) which is guaranteed, with high probability, to produce near-optimal performance against unseen switching weights of the opposition. This is empirically demonstrated in Table~\ref{table:3} which shows the superior performance of the \emph{good-for-all} policy $\mathcal{C}(\widehat{\mathbf{w}})$ to that of the \emph{good-for-one} policy $\mathcal{C}(\mathbf{w})$ when tested against opponents with unseen tactic-switching weights $\mathcal{E}(\mathbf{u}^{(1)}), \mathcal{E}(\mathbf{u}^{(2)}), \ldots, \mathcal{E}(\mathbf{u}^{(6)})$. Also, similar to the case of basic stratagem in Section~\ref{skill}, the quality of those switching policies increases and converges rapidly to the optimal value when we increase the number of optimization iterations (Fig.~\ref{fig:4}) in our stratagem fusion framework, which demonstrates the its stability.

\section{Hardware Experiments}

\begin{figure*}[t]
\begin{tabular}{ccc}
\includegraphics[width=5.7cm]{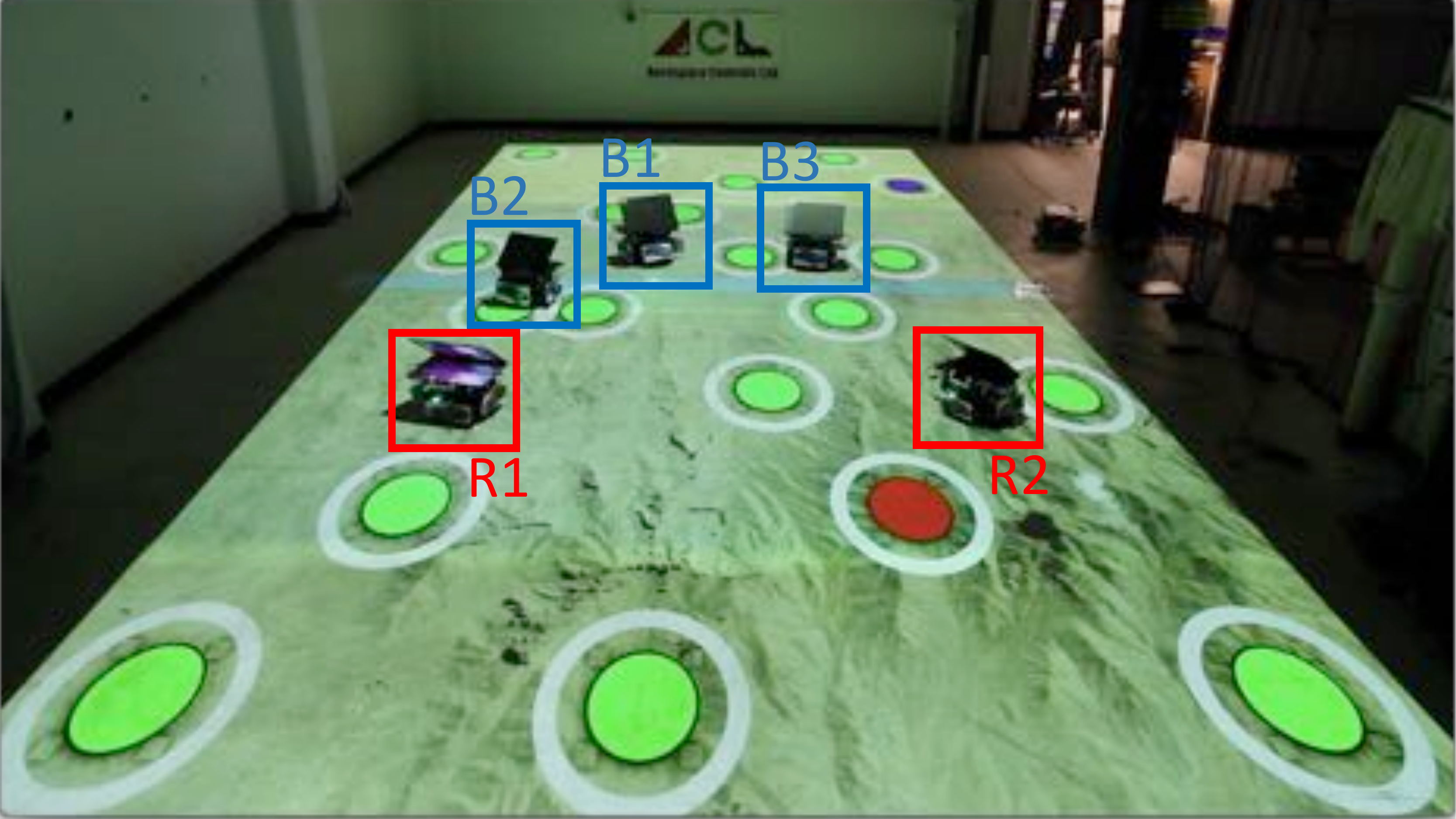} & \hspace{-2mm}\includegraphics[width=5.7cm]{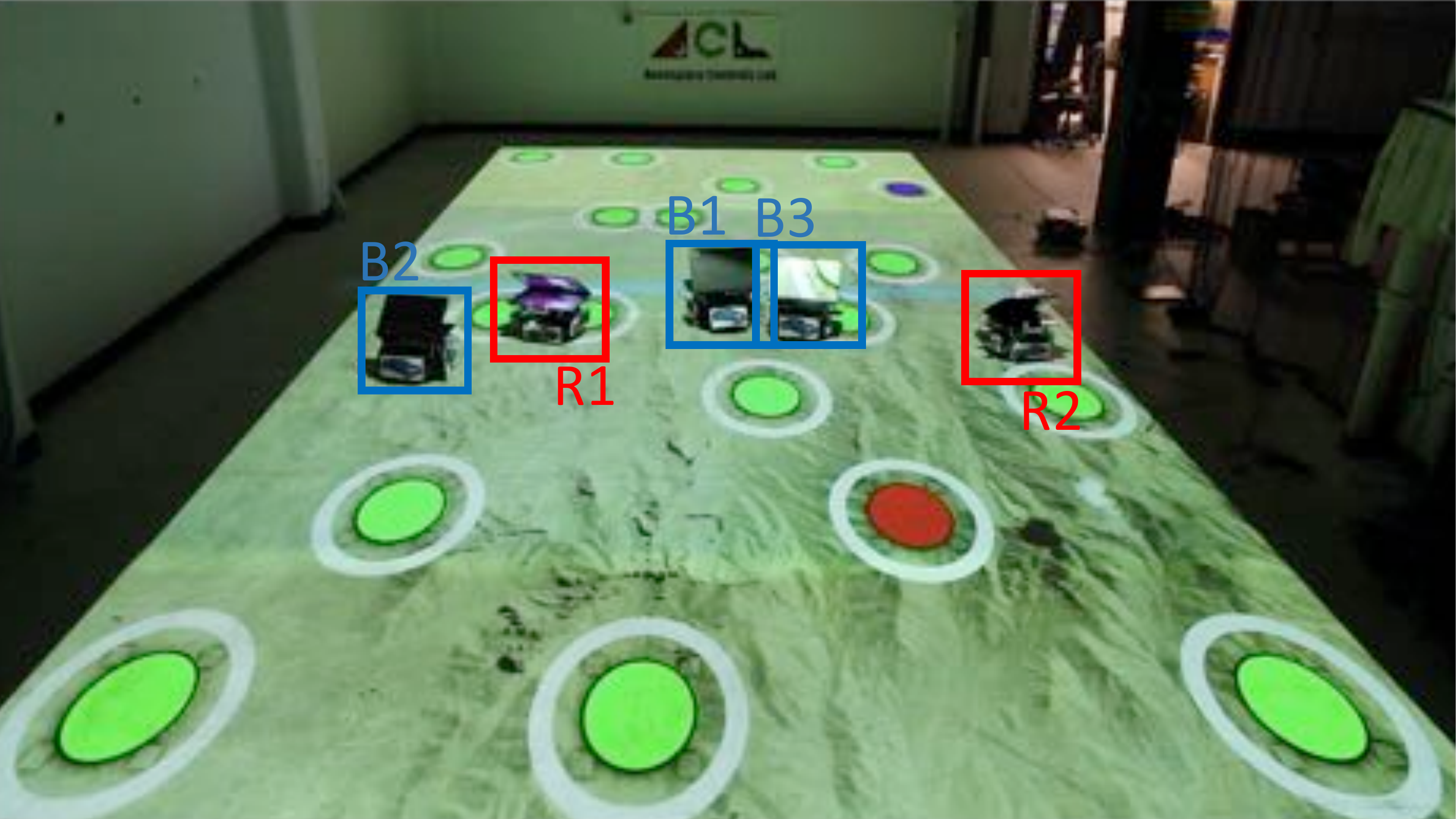} & \hspace{-2mm}\includegraphics[width=5.7cm]{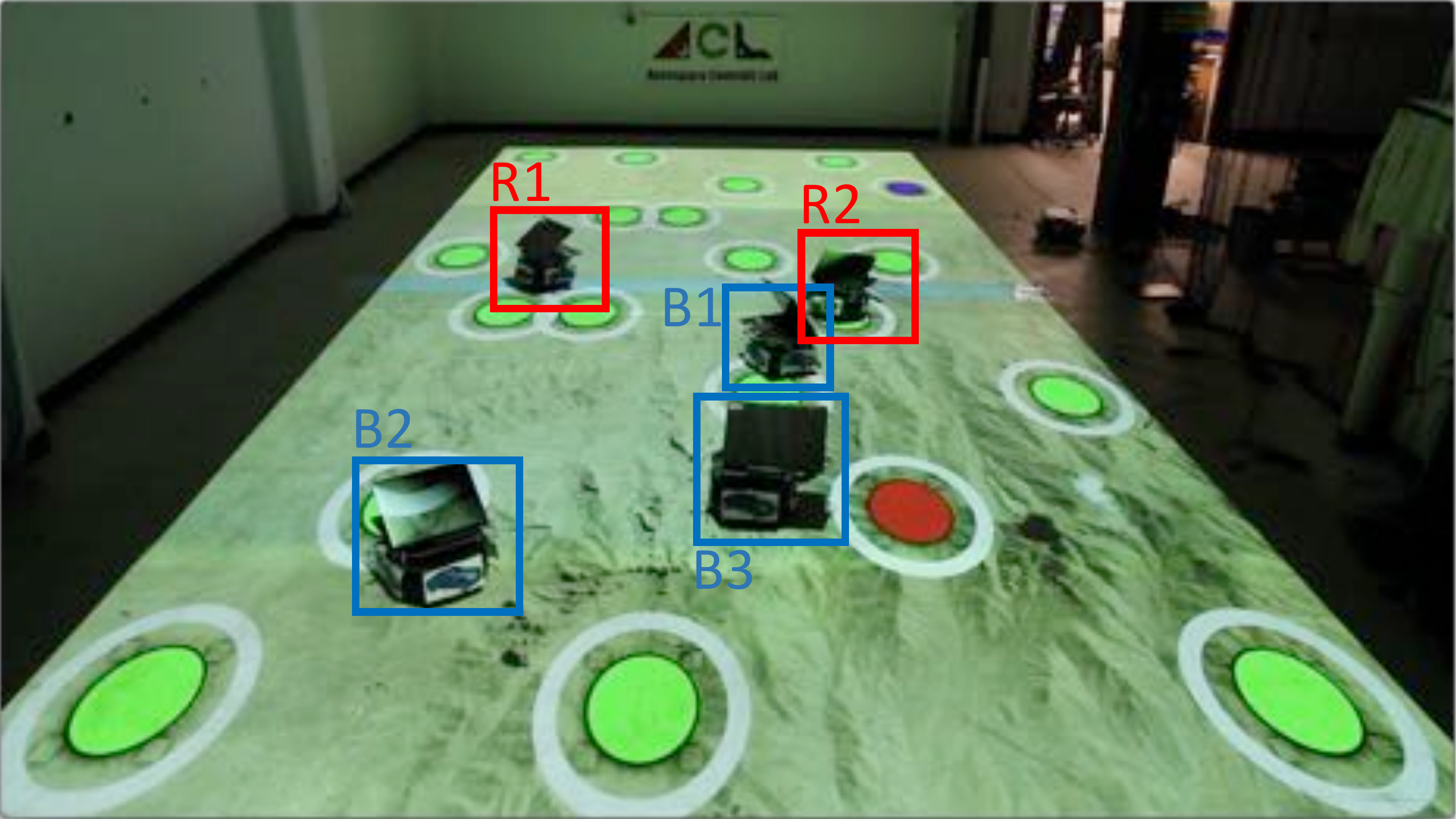}\\
(a) & (b) & (c) \\
\includegraphics[width=5.7cm]{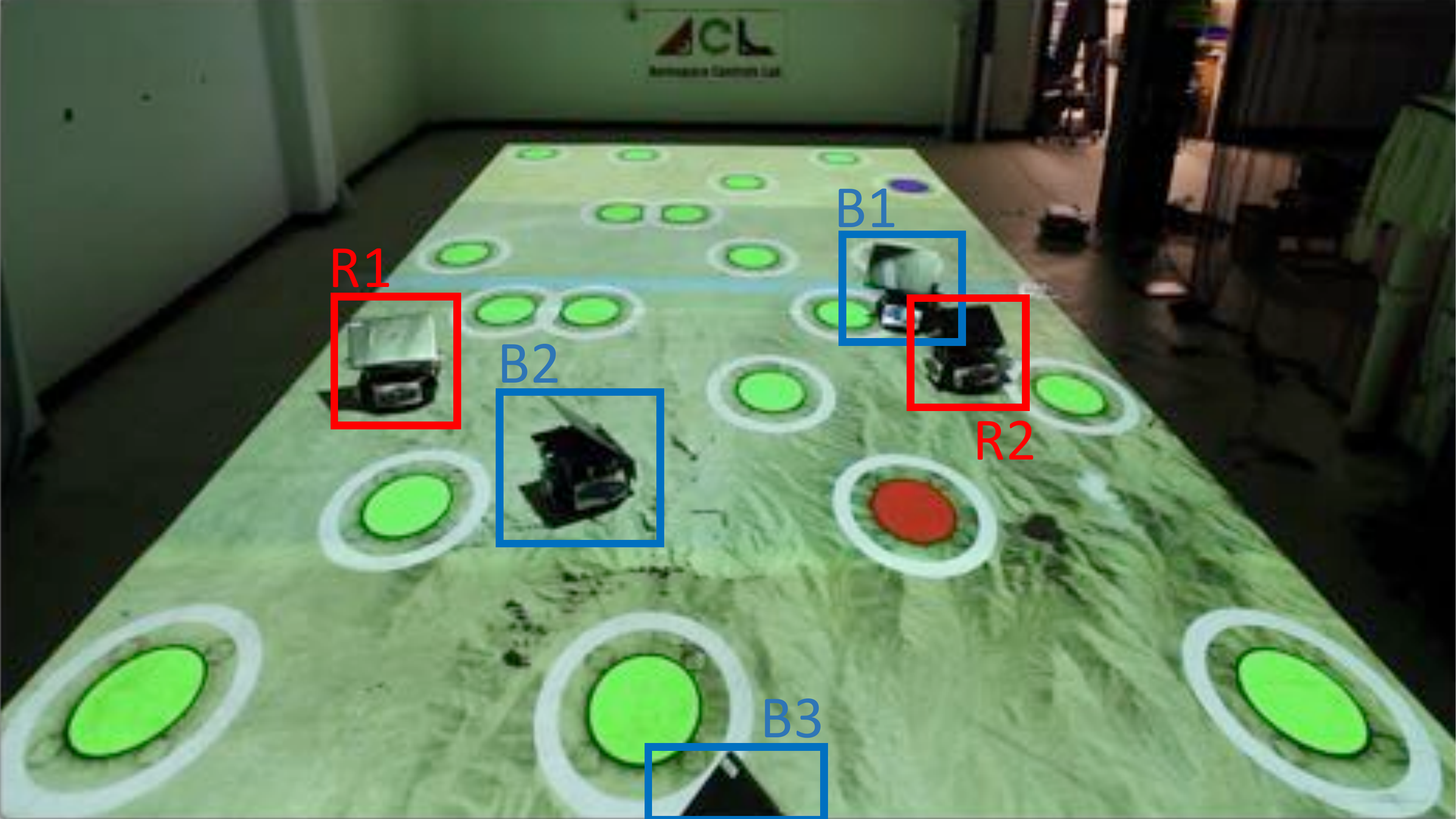} & \hspace{-2mm}\includegraphics[width=5.7cm]{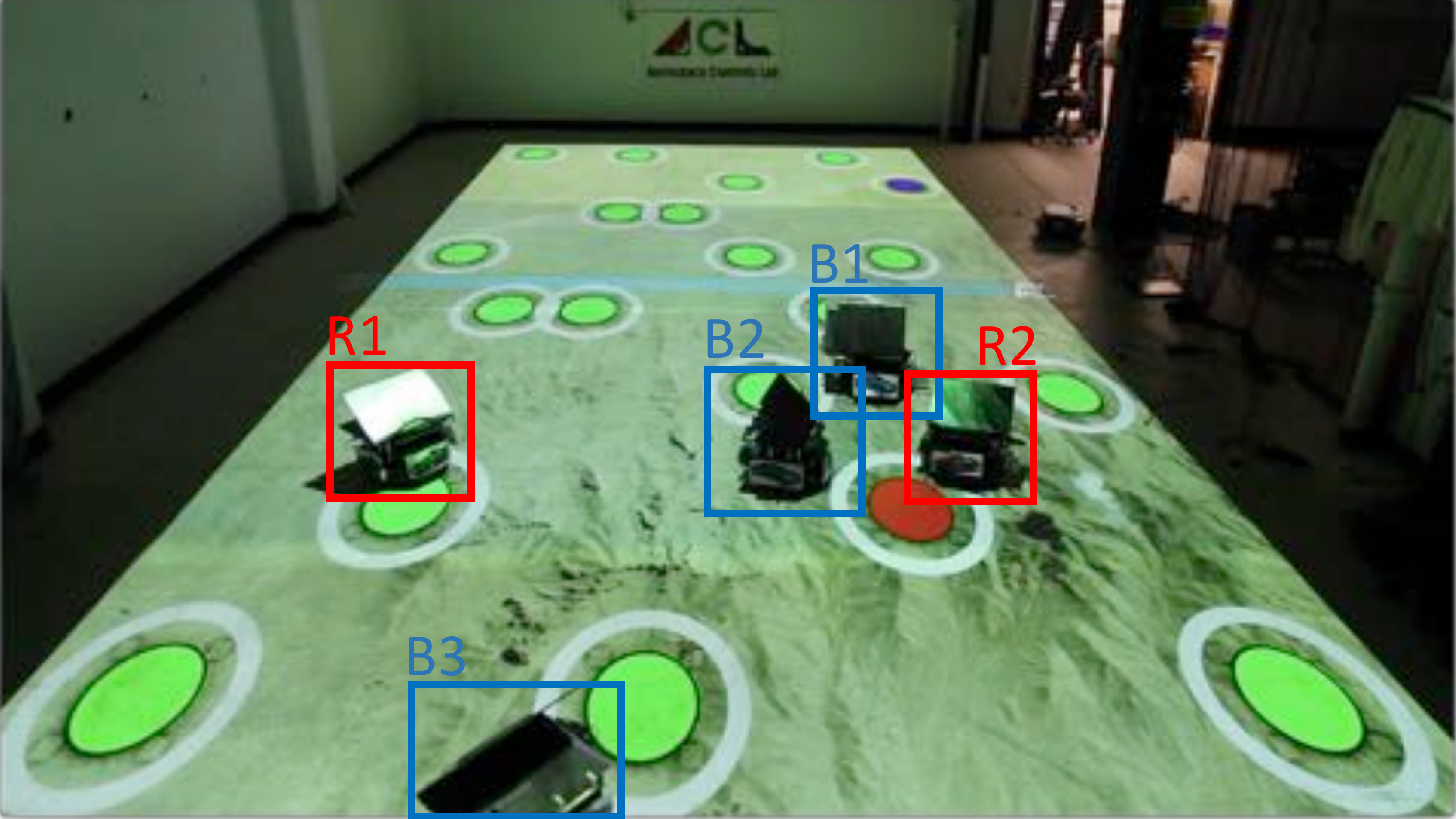} & \hspace{-2mm}\includegraphics[width=5.7cm]{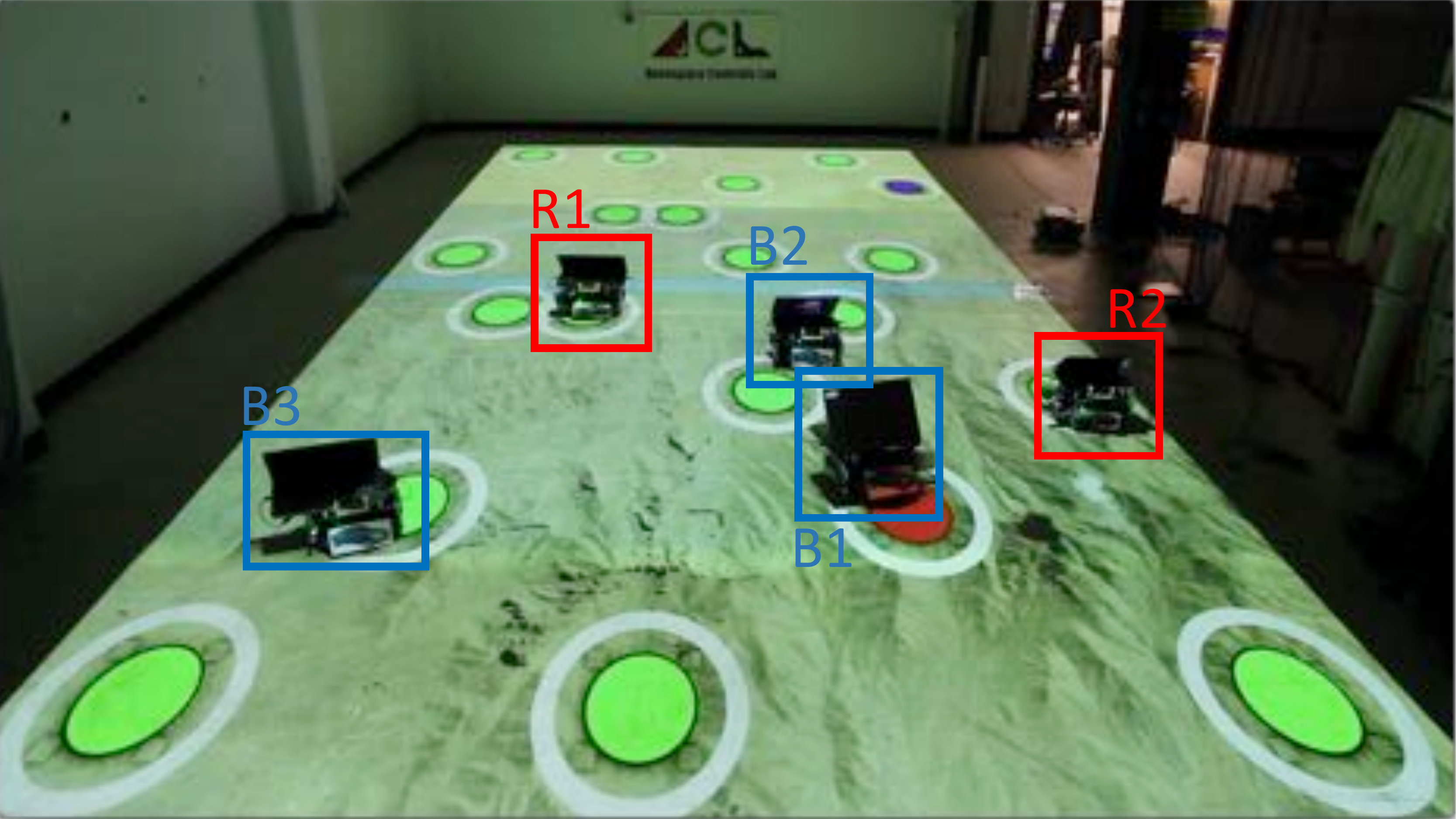}\\
(d) & (e) & (f) \\
\end{tabular}\vspace{-2mm}
\caption{Image excerpts from a video demo showing (1) a team of $3$ allied (blue) robots ($\mathbf{B1},\mathbf{B2}$ and $\mathbf{B3}$) that implement the optimized stratagem produced by our framework (Section~\ref{skill}) to compete against (2) an opposing team of $2$ opponent (red) robots ($\mathbf{R1}$ and $\mathbf{R2}$) which implement the hand-coded tactics $\mathbf{DL}$ and $\mathbf{DR}$ (see Section~\ref{CTF}), respectively: (a) $\mathbf{B1},\mathbf{B2}$ and $\mathbf{B3}$ decide to invade the opposition territory; (b) $\mathbf{B1}$ and $\mathbf{B3}$ decide to attack the center while $\mathbf{B2}$ decides to take the left flank of the opposition; (c) $\mathbf{B2}$ passes through $\mathbf{R1}$'s defense while $\mathbf{B1}$ takes an interesting position to block $\mathbf{R2}$ so that $\mathbf{B3}$ can pass through its defense; (d) $\mathbf{B1}$ and $\mathbf{B2}$ detect the flag and mount a pincer attack; (e) $\mathbf{R2}$ arrives to defend the flag and $\mathbf{B2}$ retreats to avoid getting tagged; and (f) without noticing $\mathbf{B1}$ from behind, $\mathbf{R2}$ continues its $\mathbf{DR}$ patrol, thus losing the flag to $\mathbf{B1}$.}\vspace{-7mm}
\label{fig:6}
\end{figure*}

\noindent In addition to the simulated experiments, we also conduct real-time experiments with real robots to showcase the robustness of our proposed framework in practical RTS scenarios. The specifics of our robot configuration and domain setup are shown in Fig.~\ref{fig:5}. Each robot is built with the Kobuki base of TurtleBot $2$ and configured with on-board processing unit (Gigabyte Aero 14 laptop with Intel Core i$7$-$7700$HQ quad-core CPU and NVIDIA GTX $1060$ GPU with $6$GB RAM) as well as sensory devices including (1) Intel RealSense Camera (R$200$) Developer Kit ($130$mm x $20$mm x $7$mm) with Depth/IR: Up to $640 \times 480$ resolution at $60$ FPS $\&$ RGB: $1080$p at $30$ FPS; and (2) Omnidirectional RPLIDAR A2 ($4000$ samples/sec ($10$Hz) and $8/16$m range). The information provided by the LIDAR sensor is directed to each robot's on-board collision-avoidance navigation procedure \cite{YuFan16} to helps it localize and move around without colliding with other robots and obstacles in the environment. The visual feed from RealSense camera is passed through the Single Shot MultiBox Detector \cite{Wei16} implemented on each allied robot's processing unit to detect its surrounding objects (e.g., the opposing robots, other allied robots and flags). The processed information is then used to generate the high-level macro-observations (Section~\ref{CTF}) for the robot's on-board policy controller. Fig.~\ref{fig:6} shows a visual excerpt from our video demo featuring a CTF scenario of $3$ allied robots which implement the optimized policy produced by our framework to compete against an opposing team of $2$ adversary robots implementing the hand-coded tactics in Section~\ref{CTF}. The excerpt shows interesting teamwork between all allied robots in capturing the opposing team's flag despite their partial, decentralized views of the world (see detailed narration in Fig.~\ref{fig:6}'s caption), which further demonstrates the robustness of our proposed framework in practical robotic applications. Interested readers are referred to our attached video demo for a complete visual demonstration.




\section{CONCLUSION}
\label{conclusion}
\vspace{-1mm}
This paper introduces a novel near-optimal adversarial policy switching algorithm for decentralized, non-cooperative multi-agent systems. Unlike the existing works in literature which are mostly limited to simple decision-making scenarios where a single agent plans its best response against an adversary whose strategy is specified a priori under reasonable assumptions, we investigate instead a class of multi-agent scenarios where multiple robots need to operate independently in collaboration with their teammates to act effectively against adversaries with changing strategies. To achieve this, we first optimize a set of basic stratagems that each is tuned to respond optimally to a pre-identified basic tactic of the adversaries. The stratagems are then integrated into a unified policy which performs near-optimally against any high-level strategies of the adversaries that switches between their basic tactics. The near-optimality of our proposed framework can be established in both theoretical and empirical settings with interesting and consistent results. We believe this is a significant step towards bridging the gap between theory and practice in multi-agent research.\vspace{1mm}

{\bf \noindent Acknowledgements.} This research is funded in part by the ONR under MURI program award $\#$N000141110688 and BRC award $\#$N000141712072 and Lincoln Lab.\vspace{-2mm}

\bibliographystyle{unsrt}
\bibliography{icra18}
\vspace{-2mm}

\if \myproof1	
\clearpage

\appendix
\balance

{\bf\noindent Proof of Lemma 1.} \vspace{-2mm}\\

\noindent The result of Lemma 1 follows directly from the classical PAC-Bayes result \cite{McAllester99} in learning theory. This is achieved by setting up an artificial learning task where $\mathbb{L}(\mathbf{w})$ (\eqref{eq:3}) corresponds to the expected loss of a hypothesis candidate $\mathbf{w}$ when evaluated against a random test input $\mathbf{u} \sim \pi(\mathbf{u})$ and similarly, $\widehat{\mathbb{L}}(\mathbf{w})$ (\eqref{eq:4}) denotes the average performance of $\mathbf{w}$ on a training set of sampled tests $\{\mathbf{u}^{(i)}\}_{i=1}^m$ drawn i.i.d from $\pi(\mathbf{u})$. The sampling distribution $q(\mathbf{w};\theta)$ can then be treated as a stochastic predictor that samples a hypothesis to evaluate a given test input. Its empirical and generalized loss thus corresponds to $\mathbb{L}(q)$ and $\widehat{\mathbb{L}}(q)$, respectively. Finally, pretending that $p(\mathbf{w})$ is the prior over the hypothesis space of $\mathbf{w}$, the gap between $\mathbb{L}(q)$ and $\widehat{\mathbb{L}}(q)$ can therefore be bounded with high probability using \eqref{eq:5}, which is the exact statement of the PAC-Bayes bound for stochastic predictors in \cite{McAllester99}.\\

{\noindent\bf Proof of Lemma 2.} \vspace{-2mm}\\

\noindent By Lemma 1, with probability at least $1 - \delta/2$, 
\begin{eqnarray}
\hspace{-7mm}\mathbb{L}(q) &\leq& \widehat{\mathbb{L}}(q) + \left(\frac{\mathbb{D}_{\mathrm{KL}}\left(q(\mathbf{w};\theta^\ast) \|p(\mathbf{w})\right) + \log \frac{8m}{\delta}}{2m - 1}\right)^{\frac{1}{2}} \label{eq:7} 
\end{eqnarray}
On the other hand, by Assumption 1, we also know that 
\begin{eqnarray}
\mathbb{D}_{\mathrm{KL}}\left(q(\mathbf{w};\theta^\ast) \|p(\mathbf{w})\right) &\leq& \epsilon_{h,\delta/2} \label{eq:8}
\end{eqnarray}
where $h$ is the size of $\mathbf{w}$, which contains the weights associated with the inter-controller transitions in the agents' high-level controllers. Since each agent has $r$ low-level controllers (corresponding to its $r$ stratagems) and there are $k$ nodes in each controller, there will be $r(r-1)k^2$ inter-controller transitions within each agent's high-level controller. Thus, in total, a set of high-level controllers for $n$ agent has $nr(r-1)k^2$ inter-controller transitions. The number $h$ of weights associated with those transitions are therefore $O(nr(r-1)k^2)$. Finally, plugging \eqref{eq:8} into \eqref{eq:7} yields \eqref{eq:6}. Since \eqref{eq:7} and \eqref{eq:8} hold with probability at least $1 - \delta/2$ each, \eqref{eq:6} holds with probability at least $1 - \delta$ by the union bound.\\

{\noindent\bf Proof of Lemma 3.} \\

\noindent By definition, we have $\widehat{\mathbb{L}}(q) \leq \widehat{\mathbb{L}}(p) = \widehat{\mathbb{L}}(\widehat{\mathbf{w}})$. Then, note that $\mathbb{E}[\widehat{\mathbb{L}}(\widehat{\mathbf{w}})] = \mathbb{L}(\widehat{\mathbf{w}})$ due to the definitions in \eqref{eq:3} and \eqref{eq:4}. As such, by Hoeffding inequality, it follows that with probability at least $1 - \mathrm{exp}(-2m\epsilon^2)$, 
\begin{eqnarray}
\hspace{-0mm}\widehat{\mathbb{L}}(q) \ \leq\ \widehat{\mathbb{L}}(\widehat{\mathbf{w}}) \ \leq\ \mathbb{L}(\widehat{\mathbf{w}}) \ +\  \epsilon \ \leq\ \mathbb{L}(\mathbf{w}^\ast) \ +\ \epsilon \ , \label{eq:9a}
\end{eqnarray}
where the third inequality follows directly from the definition of $\mathbf{w}^\ast$. Thus, with probability at least $1 - \mathrm{exp}(-2m\epsilon^2)$, $\widehat{\mathbb{L}}(\widehat{\mathbf{w}}) \leq \mathbb{L}(\mathbf{w}^\ast) +\epsilon$. Setting $\delta = \mathrm{exp}(-2m\epsilon^2)$ and solving for $\epsilon$ yields $\epsilon = (\log(1/\delta)/2m)^{1/2}$. Substituting this into Eq.~\eqref{eq:9a} yields \eqref{eq:9}.\\

\noindent {\bf Proof of Theorem 1.}\\

\noindent Applying Lemma 2 with $\delta/2$ yields 
\begin{eqnarray}
\hspace{-8mm}\mathbb{L}(q) &\leq& \widehat{\mathbb{L}}(q) \ +\ \left(\frac{\epsilon_{h,\frac{\delta}{4}} + \log \frac{16m}{\delta}}{2m - 1}\right)^{\frac{1}{2}} \ .\label{eq:10}
\end{eqnarray}
Likewise, applying Lemma 3 with $\delta/2$ yields
\begin{eqnarray}
\widehat{\mathbb{L}}(q) &\leq& \mathbb{L}(\mathbf{w}^\ast) \ +\ \left(\frac{\log\frac{2}{\delta}}{2m}\right)^{\frac{1}{2}} \ , \nonumber\\
&\leq& \mathbb{L}(\mathbf{w}^\ast) \ +\ \left(\frac{\epsilon_{h,\frac{\delta}{4}} + \log \frac{16m}{\delta}}{2m - 1}\right)^{\frac{1}{2}} \ .\label{eq:11}
\end{eqnarray}
Substituting \eqref{eq:11} into \eqref{eq:10} yields \eqref{eq:12}. Since \eqref{eq:10} and \eqref{eq:11} hold with probability at least $1 - \delta/2$ each, \eqref{eq:12} holds with probability at least $1 - \delta$ by the union bound. This result thus guarantees the near-optimality of our stratagem integration scheme with high probability.\vspace{-1mm}\\

{\noindent\bf Remark.} Both the PAC-Bayes result of \cite{McAllester99} and Hoeffding inequality in Lemma 3 implicitly assume the loss function is always bounded between $0$ and $1$. This is however not a hindrance to our analysis since MacDEC-POMDP's intrinsic reward function can always be re-scaled to meet that requirement without affecting the solution quality.

\end{document}